%% file: neurips_2025.tex
\documentclass{article}

 \usepackage[main, final]{neurips_2025}

\usepackage[utf8]{inputenc} 
\usepackage[T1]{fontenc}    
\usepackage{hyperref}       
\usepackage{url}            
\usepackage{booktabs}       
\usepackage{amsfonts}       
\usepackage{nicefrac}       
\usepackage{microtype}      
\usepackage{xcolor}         

\usepackage{listings}
\usepackage[most]{tcolorbox}
\usepackage{xcolor}
\usepackage{enumitem}
\usepackage{amsmath} 
\usepackage{algorithm}
\usepackage{algpseudocode}
\usepackage{marvosym}
\usepackage{xspace}
\usepackage{cleveref}
\usepackage{wrapfig}
\usepackage{amssymb}
\usepackage{float}

\definecolor{algcolor1}{RGB}{68,1,84}        
\definecolor{algcolor2}{RGB}{253,231,37}    
\definecolor{algcolor3}{RGB}{68,119,170}    
\definecolor{algcolor4}{RGB}{94,201,98}     
\definecolor{algcolor5}{RGB}{36,169,225}    
\definecolor{algcolor6}{RGB}{253,141,60}    

\definecolor{low}{RGB}{68, 1, 84} 
\definecolor{medium}{RGB}{33, 144, 141} 
\definecolor{high}{RGB}{94, 201, 98} 

\newcommand{\RepeatIndent}[1]{%
  \ifcase#1\relax
  \or \hspace{\algorithmicindent}%
  \or \hspace{\algorithmicindent}\hspace{\algorithmicindent}%
  \or \hspace{\algorithmicindent}\hspace{\algorithmicindent}\hspace{\algorithmicindent}%
  \or \hspace{\algorithmicindent}\hspace{\algorithmicindent}\hspace{\algorithmicindent}\hspace{\algorithmicindent}%
  \else \hspace{#1\algorithmicindent} 
  \fi
}
\newcommand{\AlgComment}[2]{\Statex \RepeatIndent{#1} \textnormal{// #2}}

\newcommand{\decomp}{\textsc{DISC}\xspace}

\newtheorem{theorem}{Theorem}
\newtheorem{lemma}{Lemma}
\newtheorem{assumption}{Assumption}
\newtheorem{proof}{Proof}
\newtheorem{remark}{Remark}

\title{\decomp: Dynamic Decomposition Improves LLM Inference Scaling}

\author{%
  Jonathan Light$^{1,2, 4}$\thanks{Work done during the part-time internship at NEC Laboratories America. \textsuperscript{\Letter}Corresponding author.} \thanks{This work was partially supported by IBM through the IBM-Rensselaer Future of Computing Research Collaboration.}\ \,, Wei Cheng$^2$\textsuperscript{\Letter},\ \ Benjamin Riviere$^4$,\ \ Yue Wu$^3$,\ \ Masafumi Oyamada$^5$,\\ 
  \textbf{Mengdi Wang}$^3$,\ \ \textbf{Yisong Yue}$^4$,\ \ \textbf{Santiago Paternain}$^1$,\ \ \textbf{Haifeng Chen$^2$} \\
  $^1$\textmd{Rensselaer Polytechnic Institute},\ \ 
  $^2$\textmd{NEC Laboratories America},\ \ 
  $^3$\textmd{Princeton University},\\ 
  $^4$\textmd{California Institute of Technology},\ \ 
  $^5$\textmd{NEC Corporation} \\
}

\begin{document}

\maketitle

\vspace{-3em}
\begin{center}
  \large\href{https://disc-search.github.io/}{\textcolor{blue}{\texttt{disc-search.github.io}}}
\end{center}
\vspace{1em}

\begin{abstract}
    Inference scaling methods for LLMs often rely on decomposing problems into steps (or groups of tokens), followed by sampling and selecting the best next steps. 
    However, these steps and their sizes are often predetermined or manually designed based on domain knowledge.
    We propose dynamic decomposition, a method that adaptively and automatically partitions solution and reasoning traces into manageable steps during inference.
    By more effectively allocating compute -- particularly through subdividing challenging steps and prioritizing their sampling -- dynamic decomposition significantly improves inference efficiency.
    Experiments on benchmarks such as APPS, MATH, and LiveCodeBench demonstrate that dynamic decomposition outperforms static approaches, including token-level, sentence-level, and single-step decompositions, reducing the pass@10 error rate by 5.0\%, 6.7\%, and 10.5\% respectively. These findings highlight the potential of dynamic decomposition to improve a wide range of inference scaling techniques.
    
\end{abstract}

\input{sections/introduction}
\input{sections/preliminaries}

\input{sections/methodology_2}
\input{sections/results}
\input{sections/related_works}
\input{sections/conclusion}


\clearpage
\newpage
\bibliographystyle{unsrt}
\bibliography{ref}

\clearpage
\tableofcontents
\clearpage
\appendix

\input{sections/appendix}

\end{document}

%% file: sections/introduction.tex
\section{Introduction}

Scaling inference efficiency remains a fundamental challenge for large language models (LLMs). Many existing approaches improve inference by decomposing problems into smaller steps and systematically exploring different solutions~\citep{feng2023alphazero, zeng2024scaling, wu2024inferencescalinglawsempirical, nori2024medprompto1explorationruntime, snell2024scaling, brown2024large, gandhi2024stream, lee2025evolvingdeeperllmthinking, light2024strategist, anonymous2025planning}. 
Some decomposition methods often rely on domain-specific heuristics and hand-crafted rules~\citep{yao2024tree, zelikman2023parsel, zhou2022least}. However, manually partitioning problems or designing task-specific heuristics is costly and lacks generalization. 
Moreover, identifying critical steps  for an LLM can be non-trivial for humans. LLMs may assign importance to seemingly trivial words (e.g., therefore or which), which, while counterintuitive to humans, play a crucial role in autoregressive generation \citep{lin2025criticaltokens}.
Other approaches employ fixed, uniform step sizes, such as token- or sentence-level decomposition~\citep{feng2023alphazero, guo2025deepseek}. 
All these methods rely on \textbf{static decomposition strategies}, where step sizes are predefined or determined via heuristics. 
Such rigidity risks overusing compute on steps that are easy for the LLM (but potentially difficult for humans) while undersampling more challenging steps.

To overcome these limitations, we propose \decomp (\underline{D}ynamic decomposition \underline{I}mproves \underline{S}caling \underline{C}ompute), a recursive inference algorithm that dynamically partitions solution steps based on difficulty. Unlike prior methods, \decomp \textbf{adapts decomposition granularity} during inference based on both the available budget and problem complexity, ensuring finer granularity for more difficult steps. By leveraging the autoregressive nature of LLMs, \decomp efficiently \textbf{locates difficult steps} through dynamically proposing step sizes, focusing compute on challenging regions rather than wasting resources on trivial steps. \decomp is generalizable, requires \emph{no human supervision, domain-specific heuristics, prompt engineering, or process annotations}, and is easily \emph{parallelizable}, making it widely applicable across tasks. Furthermore, \decomp is plug-and-play with off-the-shelf search algorithms and can be naturally integrated with greedy, beam, or Monte Carlo Tree Search.

Our main contributions are:
\begin{itemize}[leftmargin=*,itemsep=0pt,topsep=0pt,parsep=0.4pt, partopsep=0pt]
    \item We introduce \textbf{\decomp}, a method for dynamically adjusting step sizes and decomposing solutions during inference without human supervision, domain-specific heuristics, or process reward models.

    \item We demonstrate how \decomp integrates decomposition directly into inference-time search, \textbf{allocating compute more effectively toward high-potential solution prefixes}.

    \item We show that \decomp improves inference scaling in terms of both \textbf{sample efficiency}, \textbf{token efficiency}, and \textbf{runtime}, achieving up to \textbf{10\% reduction} in error relative to the baselines and up to \textbf{4x increase} in accuracy over the base model, with just 10 samples, including with reasoning models. 

    \item We provide both empirical and theoretical insights into LLM reasoning, including identifying \textbf{critical intermediate steps} and analysis of how \decomp helps discover optimal solutions.
\end{itemize}

%% file: sections/preliminaries.tex
\section{Preliminaries}
\begin{figure*}
    \centering
    \includegraphics[width=0.95\textwidth]{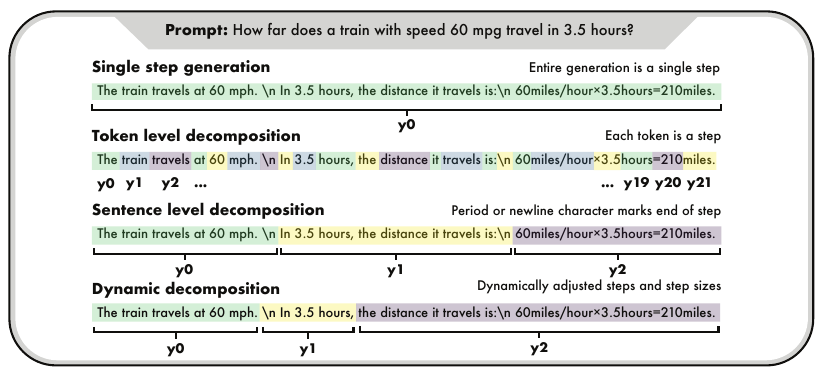}
    \caption{
    \textbf{Comparison of automatic decomposition strategies based on step size.}  
    Coarser steps accelerate the search process but risk skipping over optimal solutions and committing to suboptimal prefixes. In contrast, finer steps ensure more precise decisions but lead to slower search. A dynamic strategy that adapts step size based on LLM feedback offers a balanced approach, combining the efficiency of coarse steps with the precision of fine-grained decomposition.
    }
    \label{fig:decomp_comparison}
\end{figure*}

\subsection{Problem Setting}
\vspace{-0.1in}
We consider a reasoning and code generation setting where we are given: 
a dataset $\mathcal{X} = \{\boldsymbol{x}^{(i)}\}_{i=1}^{N}$, 
a pretrained, autoregressive LLM $\pi$ that generates solutions $\boldsymbol{y} \in \mathcal{Y}$, 
and a reward model $R: \mathcal{X} \cdot \mathcal{Y} \rightarrow [0,1]$ that evaluates the generated solutions. 
The goal is to find inputs to the LLM that produce solutions with high reward. 
This setting includes program synthesis, where correctness is verified using ground-truth tests~\citep{chen2021evaluating, austin2021program}, and mathematical reasoning, where solutions are validated numerically~\citep{hendrycks2021measuring, cobbe2021training}. 
The reward model can be a ground-truth verifier, a trained heuristic~\citep{zhang2024generativeverifiersrewardmodeling}, self-consistency~\citep{wang2023selfconsistency}, or an LLM-as-a-judge~\citep{zheng2023judgingllmasajudgemtbenchchatbot} and because our focus is on decomposition rather than verification, we use the ground-truth reward model where available.

We use the following hierarchical token notation: 
$y$ is a token, 
$\boldsymbol{p}$ is a prefix that starts at the prompt and concatenates multiple steps, $\boldsymbol{s}$ is a suffix that concatenates multiple steps and ends with the \textsc{EOS} token, and 
$\boldsymbol{y}$ is a complete solution that starts at the prompt, concatenates multiple steps, and ends at the end of sequence token \textsc{EOS}. 
We denote a concatenation of two tokens or token sequences with $\cdot$, e.g. $\boldsymbol{p} \cdot \boldsymbol{s}$ is the concatenation of prefix $\boldsymbol{p}$ and suffix $\boldsymbol{s}$ to form a complete solution $\boldsymbol{y}$.
We denote the sampled suffix from prefix $\boldsymbol{p}$ using the LLM policy as: $\boldsymbol{s} \sim \pi(\cdot | \boldsymbol{p})$. 
We denote an optimal solution with $\boldsymbol{y}^*$ and an optimal suffix as $\boldsymbol{s}^*$, where there exist multiple optimal solutions. 
For our analysis, we use the convention that $\pi(\boldsymbol{s}^*|\boldsymbol{p}) = 0$ if there does not exist a completion $\boldsymbol{s}^*$ such that $\boldsymbol{p} \cdot \boldsymbol{s}^*$ is optimal. 
The \textbf{size} of a string $|\boldsymbol{y}|$, refers to its length in tokens or characters.

\subsection{Existing Decomposition Methods}
\label{sec:prior_decomp}
\textbf{Single-step generation.}  
In a single-step generation, the entire solution is generated in one pass from the prompt to the \textsc{EOS} token, treating it as a single action. This approach underlies the widely used inference scaling method \textbf{best of n} (BoN)~\citep{cobbe2021training, lightman2023let, snell2024scaling, liang2024improving}, where $n$ complete solutions are sampled, and the highest-scoring one is selected. Single-step generation also plays a role in alignment and fine-tuning methods such as \textbf{DPO}~\citep{rafailov2024direct} and \textbf{RLOO}~\citep{ahmadian2024back}.

\textbf{Token-level decomposition.}  
At the opposite end of the spectrum, token-level decomposition treats each atomic token as an individual step. While this approach dramatically increases search complexity, it enables fine-grained search that can yield higher performance gains given sufficient compute~\citep{feng2023alphazero}.

\textbf{Newline and sentence-level decomposition.}  
A commonly used decomposition method segments LLM generations into sentences or lines based on delimiters such as periods or newlines~\citep{hao2023reasoning, feng2023alphazero, yao2024tree}. Typically, each newline corresponds to a new paragraph, equation, or line of code, which often encapsulates a distinct reasoning step.

\begin{tcolorbox}[title=Challenge: Automatic and scalable decomposition, colframe=low]
    Existing decomposition methods are static and manually designed, resulting in either slow convergence to good performance or fast convergence to poor performance. We propose adaptive decomposition for fast convergence to a good performance. 
\end{tcolorbox}


\vspace{-0.1in}
\subsection{Search for Inference Scaling}
\vspace{-0.1in}
We differentiate between decomposition and search, where decomposition controls the number of steps between new branches and search is the process of selecting which branch to explore. 
In other words, decomposition is the construction of nodes and edges, and search is a process that occurs on that structure. 
In this work, we propose a decomposition method that is plug-and-play with different search methods. 
In our experimental results in Sec.~\ref{sec:scaling_methods}, we compare decomposition methods: token-level decomposition, sentence-level decomposition, and \decomp with the search methods: greedy, beam~\citep{xie2024self}, and Monte Carlo Tree Search~\citep{feng2023alphazero, light2024scattered}. 
Implementation details are provided in App.~\ref{sec:search_extended}.

%% file: sections/methodology_2.tex
\begin{wrapfigure}{r}{0.4\textwidth} 
    \vspace{-0.4in}
    \centering
    \includegraphics[width=\linewidth]{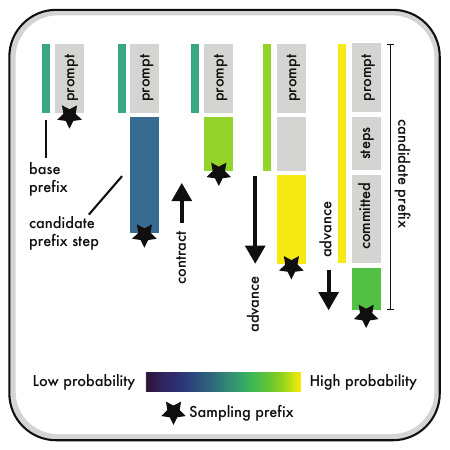}
    \vspace{-0.6cm}
    \caption{\footnotesize \textbf{Multiple iterations of Alg.~\ref{alg:discwithgreedy}.} \decomp dynamically refines its step sizes across iterations, advancing and contracting the prefix at which it samples from.\normalsize}
    \vspace{-0.3in}
    \label{fig:recursive_partition}
\end{wrapfigure}

\section{Methodology}

The \decomp algorithm uses LLM rollout data to dynamically decompose reasoning steps, adjust step sizes, and allocate sampling compute. 
In Sec. ~\ref{sec:disc_with_greedy}, we present \decomp paired with a greedy search strategy, shown in Alg.~\ref{alg:discwithgreedy} and Fig.~\ref{fig:tree_comp}. 
In Sec.~\ref{sec:disc_for_search}, we present the general case of \decomp as a nodal expansion operator which is used for pairing with beam search and MCTS. 

\subsection{High-level Overview}

At a high level, \decomp with Greedy Search advances a base prefix $\boldsymbol{p}_b$ forward by iteratively concatenating promising steps to it. The core intuition is to dynamically allocate compute by adjusting the step size: if a prefix shows promising reward improvement, we take a large step forward; if not, we contract the step and concentrate sampling around that prefix. This adaptive behavior focuses the LLM’s effort on regions of the search space that are more likely to yield high-reward completions.

\begin{figure*}
    \vspace{-0.3cm}
    \centering
    \includegraphics[width=0.95\textwidth]{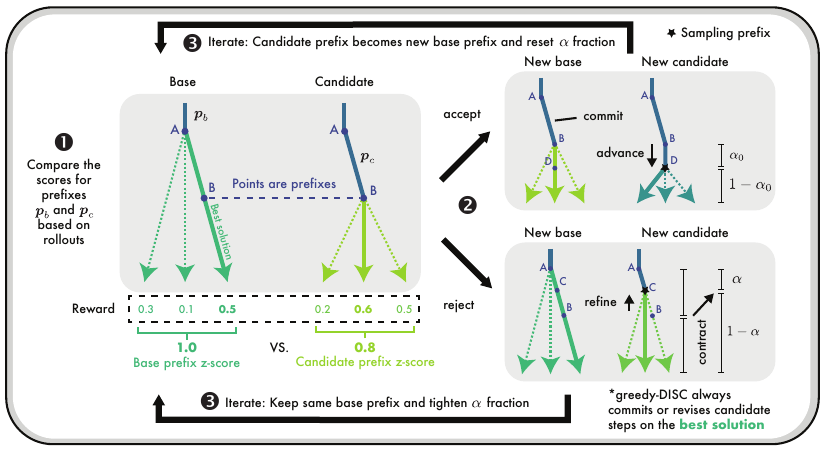}
    \caption{\textbf{\decomp with Greedy Search.} 
    One iteration of Alg.~\ref{alg:discwithgreedy}. We start with a base prefix A and a candidate prefix B. We compare the sample statistics of each by evaluating a scoring function (e.g., z-score). If the candidate prefix B demonstrates a higher likelihood of reward improvement compared to continuing from base prefix A, we accept B, commit it as the new base, and extend the candidate to a further step (e.g., BD) on the best sampled solution. If B is not better, we reject it and propose a shorter candidate (e.g., AC), contracting the step size. This process repeats until a new candidate is accepted or all options are exhausted. The algorithm thus adaptively advances or contracts the step size and search horizon based on the relative quality of completions from each prefix.
    }
    \label{fig:tree_comp}
    \vspace{-0.1in}
\end{figure*}

Fig.\ref{fig:tree_comp} illustrates a single iteration of this decision process, where a candidate prefix is either accepted and extended or rejected and contracted. Over multiple iterations, this process yields a full solution composed of several such accepted steps. Fig.\ref{fig:recursive_partition} shows how the prefix is incrementally constructed: it displays the number of steps \decomp has committed to the prefix, and how the prefix used for sampling dynamically changes over time. Together, these figures highlight the local step-wise decision-making and the global trajectory of prefix refinement throughout the search.

\vspace{-0.1in}
\paragraph{Assumptions.}
\decomp makes minimal, broadly applicable assumptions, enabling generality and ease of deployment. It avoids handcrafted prompts, process-level reward models, and domain-specific heuristics. Its only requirement is access to a scalar outcome reward model (ORM) to guide search. In the absence of a ground-truth ORM, self-supervised signals—like LLM critiques or unit tests—serve as effective substitutes~\citep{chen2022codet, mcaleese2024llm, gu2024survey}. It also assumes the underlying policy \(\pi\) can generate continuations from any prefix \(\boldsymbol{p}\), a standard feature of decoder-only language models.

\subsection{\decomp with Greedy Search Algorithm}
\label{sec:disc_with_greedy}
\begin{wrapfigure}{r}{0.4\textwidth} 
    \vspace{-0.4in}
    \centering
    \includegraphics[width=\linewidth]{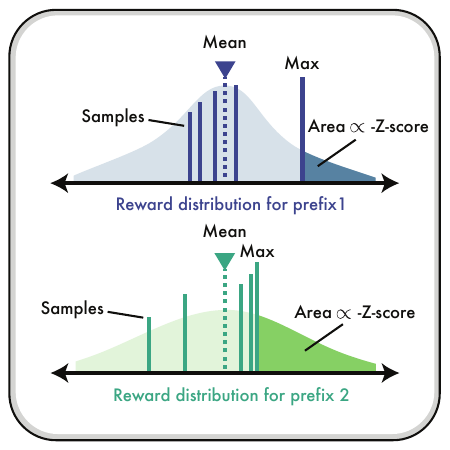}
    \vspace{-0.2in}
    \caption{\footnotesize Reward distribution of rollouts sampled from two different prefixes. The probability of sampling a higher rollout from prefix 2 is higher than that of prefix 1. \normalsize}
    \vspace{-0.3in}
    \label{fig:reward_dist}
\end{wrapfigure}

We now describe Alg.~\ref{alg:discwithgreedy}. The algorithm takes as input an LLM $\pi$, a reward model $R$, prompt $\boldsymbol{x}$, initial partition fraction $\alpha_0$, threshold $\sigma$, and sample budget $N$. It initializes the base prefix as $\boldsymbol{p}_b = \boldsymbol{x}$ and sets $\alpha = \alpha_0$.

At each iteration, the algorithm samples $\pi$ from $\boldsymbol{p}_b$ to generate completions $\boldsymbol{y}^i = \boldsymbol{p}_b \cdot \boldsymbol{s}^i$, computing rewards $R(\boldsymbol{y}^i)$. Sampling stops when the cumulative reward exceeds $\sigma$: $M = \min \{m \in \mathbb{Z}_{>0} \mid \sum_{i=1}^{m} R(\boldsymbol{y}^i) \geq \sigma \}$. This balances sample quantity and quality. The solutions and their rewards are stored in $Y$, and the best suffix $\boldsymbol{s}^*$ and z-score are computed.

A candidate prefix $\boldsymbol{p}_c$ is then formed by appending the first $\alpha$ fraction (token-wise) of $\boldsymbol{s}^*$ to $\boldsymbol{p}_b$. It is accepted if its reward z-score $z_c$ is lower than the current $z_b$. Assuming rewards follow a location-scale family distribution (e.g., Gaussian), a lower z-score implies a higher probability of improvement, since $\Pr(R > \text{max}_{\boldsymbol{s} \in S_{b,c}}(R(\boldsymbol{s})) = 1 - \mathrm{CDF}(z_{b,c})$ where $\mathrm{CDF}$ is the cumulative distribution function. Fig. ~\ref{fig:reward_dist} displays how we can estimate the probability of improvement from a reward distribution. A location-scale distribution of rewards is supported empirically (see App. ~\ref{sec:normal_reward_distribution}).

If the candidate is accepted, the algorithm updates the base prefix $\boldsymbol{p}_b \leftarrow \boldsymbol{p}_c$, resets $\alpha = \alpha_0$, and updates the base z-score and best suffix. If rejected, the partition fraction contracts: $\alpha \leftarrow \alpha \cdot \alpha_0$. This contraction implements \decomp's recursive refinement mechanism, focusing the search on more promising regions. The process repeats until the sample budget is exhausted or a correct solution is found.

\begin{wrapfigure}{r}{0.5\textwidth}
\vspace{-0.6in} 
\begin{minipage}{0.5\textwidth}
\begin{algorithm}[H]
\caption{\decomp with Greedy Search}
\label{alg:discwithgreedy}
\begin{algorithmic}[1]
\Require LLM $\pi$, Reward model $R$, prompt $x$, initial partition fraction $\alpha_0$, negative binomial threshold $\sigma$, sample budget $N$
\AlgComment{0}{initialize base prefix and current partition fraction}
\State $\boldsymbol{p}_b = \boldsymbol{x}$, $\alpha = \alpha_0$, $n=0$
\AlgComment{0}{compute base prefix statistics (\texttt{\string_b})}
\State $S_b = \{ \boldsymbol{p}_b \cdot \boldsymbol{s}^i\mid \boldsymbol{s}^i \sim \pi(\cdot | \boldsymbol{p}_b) \}_{i=1}^{M}$ 
\State $z_b = \frac{\text{max}_{\boldsymbol{s} \in S_b} (R(\boldsymbol{s})) - \text{mean}_{\boldsymbol{s} \in S_b} (R(\boldsymbol{s}))}{ \text{std}_{\boldsymbol{s} \in S_b} (R(\boldsymbol{s}))}$
\State $\boldsymbol{p}_b \cdot \boldsymbol{s}^*_b= \text{argmax}_{\boldsymbol{s}\in S_b} R(\boldsymbol{s})$
\While{$n < N$} 
    \AlgComment{1}{get candidate prefix (\texttt{\string_c})}
    \State $\boldsymbol{p}_c = \boldsymbol{p}_b \cdot \textup{split}(\boldsymbol{s}_b^*, \alpha)$
    \AlgComment{1}{sample and compute candidate prefix statistics (\texttt{\string_c})}
    \State $S_c = \{ \boldsymbol{p}_c \cdot \boldsymbol{s}^i \mid \boldsymbol{s}^i \sim \pi(\cdot | \boldsymbol{p}_c) \}_{i=1}^{M}$ 
    \State $z_c = \frac{\text{max}_{\boldsymbol{s} \in S_c} (R(\boldsymbol{s})) - \text{mean}_{\boldsymbol{s} \in S_c} (R(\boldsymbol{s}))}{ \text{std}_{\boldsymbol{s} \in S_c} (R(\boldsymbol{s}))}$
    \State $\boldsymbol{p}_c \cdot \boldsymbol{s}^*_c = \text{argmax}_{\boldsymbol{s}\in S_c} R(\boldsymbol{s})$
    \State $n \leftarrow n+M$
    \AlgComment{1}{accept or reject the candidate prefix}
    \If{$z_c < z_b$ or $|\boldsymbol{s}_b^*| \le 1$}
        \State $\boldsymbol{p}_b \leftarrow \boldsymbol{p}_c$, $\boldsymbol{s}^*_b \leftarrow \boldsymbol{s}^*_c$, 
        \State $\alpha \leftarrow \alpha_0$, $z_b \leftarrow z_c$
    \Else
        \State $\alpha \leftarrow \alpha_0 \alpha$ \; 
    \EndIf
\EndWhile 
\State \textbf{yield } $\boldsymbol{p}_b \cdot \boldsymbol{s}^*_b$
\end{algorithmic}
\end{algorithm}
\end{minipage}
\vspace{-0.4in}
\end{wrapfigure}

\subsection{Analysis of \decomp with Greedy Search}

\decomp exhibits two important properties: (i) the z-score decreases monotonically over the course of algorithm iterations, and (ii) the best candidate solution always has a higher reward than the best base solution. 
We leverage these properties, together with assumptions about the quality of $\pi$, to establish the following result. We also develop a motivating theoretical example in App. ~\ref{sec:motivating_example}.

\begin{theorem}[Optimality of \decomp]
\label{thm:optimality}
Consider Alg.~\ref{alg:discwithgreedy}. Suppose that for some problem $\boldsymbol{x}$, the optimal solution is in the support of $\pi(\cdot \mid \boldsymbol{x})$.
Then at some $n > 0$, the base prefix contains \textsc{EOS} token, the algorithm terminates, and this solution is an optimal solution. See App. ~\ref{sec:theory} for proof. 
\end{theorem}



\begin{remark}
Our analysis is dependent on a strong policy assumption, which is "reverse engineered" to be the weakest assumption on the policy such that our algorithm terminates at optimality.
In other words, this assumption depends on an instance-dependent property that must be checked empirically.
The purpose of our analysis is not to provide a universal guarantee, but rather to understand how adaptive decomposition method can control real-time inference compute without sacrificing optimality. 
\end{remark}


\subsection{\decomp for Plug-and-Play with Search Algorithms}
\label{sec:disc_for_search}

In the previous section, we presented \decomp with greedy search as a complete algorithm. 
However, the core of \decomp is the decomposition that controls from which token prefixes to query the model and dynamically adjusts step sizes, which can be plug-and-played with other search algorithms like Beam search and MCTS. 
The \decomp \textsc{ExpandNode} and \textsc{MakeChildren} are presented in Alg.~\ref{alg:disc_expansion}.
These methods are conceptually similar to the \decomp with greedy search, except that in the general search context, each node stores its own base prefix, and so when a node is expanded it computes its candidate prefix from the suffix to that node instead of from the best existing suffix. 
During the expansion operator, multiple sets of children are generated, but only the children from the final prefix are kept for search.

\subsection{Example Decomposition}
\label{sec:example_decomp}

Using a sampling budget of \(N = 100\) LLM calls, the decomposition of a representative MATH problem is shown below. Each step is enclosed in brackets and color-coded based on the z-score of sampling a better suffix conditioned on the current prefix. Brighter colors indicate lower z-scores, signaling higher importance.

\begin{tcolorbox}[colback=white, title=\decomp example decomposition, colframe=gray]
\textcolor{high}{ \string[ Let the length of the rectangle be \( l \) \string] } 
\textcolor{medium}{\string[ and the width of the rectangle be \( w \). Since the perimeter of the rectangle is 24 inches, we have that \( 2l + 2w = 24 \), so \( l + w = 12 \). We wish to maximize the area of the rectangle, \string]}
\textcolor{high}{\string[ which \string]}
    \textcolor{low}{\string[ is \( A = lw \). Let \( l = 12 - w \) and plug into the area:}
    \[
    \textcolor{low}{A = (12 - w)w \Rightarrow \qquad A = 12w - w^2}
    \]
    \textcolor{low}{Now, we differentiate \( A \) with respect to \( w \):}
    \[
    \textcolor{low}{A'(w) = 12 - 2w}
    \]
    \textcolor{low}{We wish to maximize \( A \), so we set \( A'(w) = 0 \), and solve for \( w \):}
    \[
    \textcolor{low}{12 - 2w = 0 \Rightarrow \qquad w = 6}
    \]
    \textcolor{low}{Since \( l = 12 - w \), we have that \( l = 12 - 6 = 6 \). Therefore, the area of the rectangle is \( A = lw = 6 \cdot 6 = \boxed{36} \). \string]}
\end{tcolorbox}
\decomp identifies the \textbf{first step} as highly important, which aligns with intuition—early reasoning forms the foundation for all subsequent steps. In contrast, the \textbf{final step}, although large, is marked as low-importance, indicating that \decomp allocated minimal compute toward refining it. This suggests that once earlier reasoning is fixed, there is limited opportunity for improvement in the final conclusion using additional sampling. Interestingly, the \textbf{third step}, beginning with \textit{"which"}, is assigned high importance and receives substantial attention from \decomp. This step appears to act as a pivotal decision point that shapes the direction of the remaining solution. This observation supports the idea that certain tokens or sub-sequences function as critical reasoning forks or pivots—consistent with findings from prior work~\citep{bigelow2024forking, muennighoff2025s1}.


\begin{tcolorbox}[title=Autoregressive models require autoregressive decomposition, colframe=low,]
    While transition words such as `which', `therefore', `wait', etc. may not appear significant to human readers, 
    our decomposition frequently identifies them as critical decision points for autoregressive LLMs trained on next-token prediction, where selecting a different token at these junctures can substantially alter the downstream reasoning and final outcome. 
    Therefore, it is essential for inference algorithms to allocate more compute toward sampling at these steps, and to identify these steps automatically through LLM data rather than through human design.
\end{tcolorbox}

%% file: sections/results.tex
\section{Experimental Results}
\subsection{Main Results}
\label{sec:scaling_methods}
\paragraph{Benchmarks.} We evaluate DISC on three benchmarks: \textbf{APPS}, \textbf{MATH}, and \textbf{LiveCodeBench}, to assess its impact on inference scaling for both coding and reasoning. 
\textbf{APPS}~\citep{hendrycks2021measuring} consists of 5000 competitive programming problems across three difficulty levels, with the competition-level subset being the hardest. We evaluate on a 200-problem subset due to computational constraints.
\textbf{MATH}~\citep{hendrycks2021measuring-MATH} comprises 12,500 math problems. Since the ground-truth verifier provides only binary rewards, we use a pretrained ORM~\citep{xiong2024rlhflowmath}, trained via the method in~\citep{wang2024math}, with Llama-3.1-8B-Instruct as the base model. We test on a 500-problem subset (\textbf{MATH500}), identical to prior work~\citep{wang2024math, lightman2023let}.
\textbf{LiveCodeBench}~\citep{jain2024livecodebench} is a continuously updated dataset from Leetcode, AtCoder, and CodeForces, ensuring LLMs have not been exposed to test problems. We evaluate on the 108 problems uploaded between 10/01/2024 and 12/01/2024 to prevent contamination.

\begin{figure}[ht]
    \centering
    \begin{minipage}{0.32\textwidth}
        \centering
        \includegraphics[width=\linewidth]{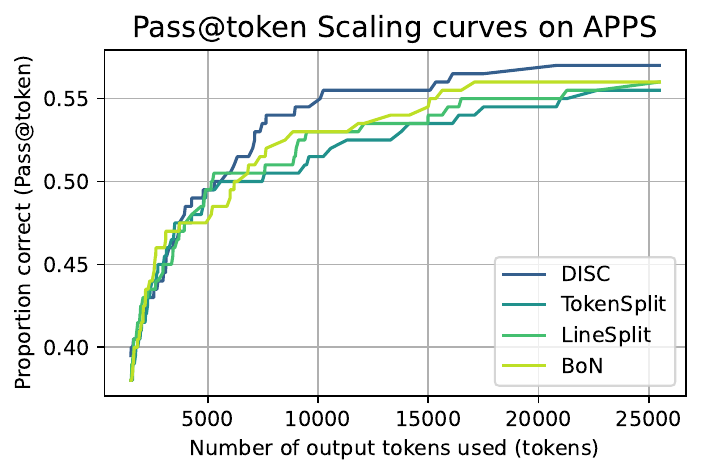}
    \end{minipage}
    \hfill
    \begin{minipage}{0.32\textwidth}
        \centering
        \includegraphics[width=\linewidth]{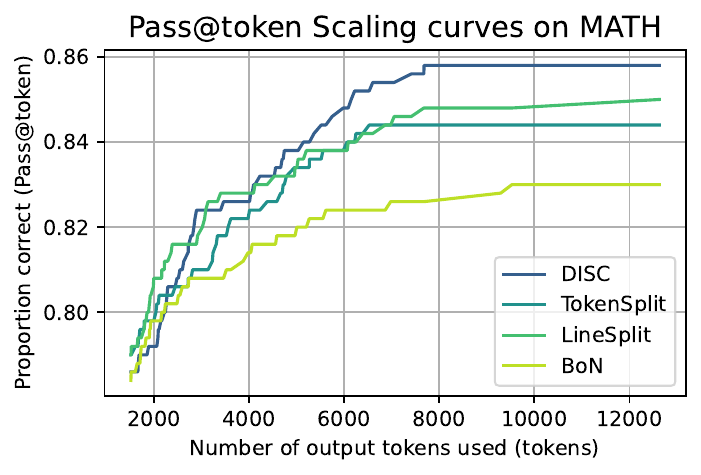}
    \end{minipage}
    \hfill
    \begin{minipage}{0.32\textwidth}
        \centering
        \includegraphics[width=\linewidth]{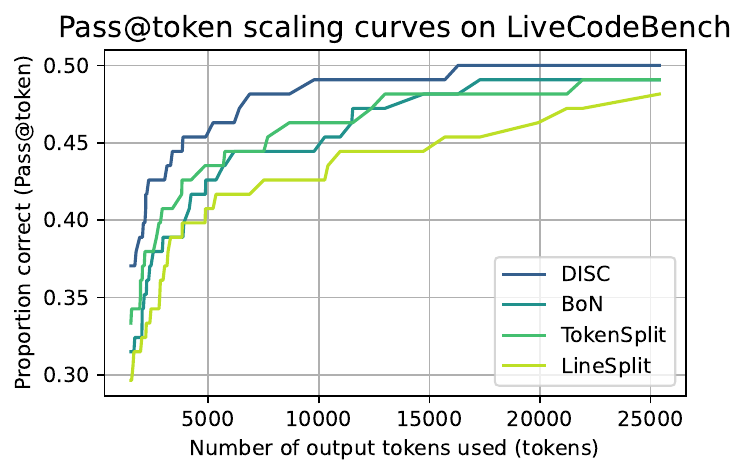}
    \end{minipage}
    \caption{\textbf{Token-level comparisons across benchmarks} using gpt-4o-mini. (Left) APPS competition level (Middle) MATH500 (Right) LiveCodeBench. \decomp achieves superior inference scaling over baselines on all three benchmarks.}
    \vspace{-0.3cm}
    \label{fig:token_comparison}
\end{figure}


\paragraph{Baselines.} We compare \decomp against three prior decomposition methods: \textbf{TokenSplit} (token-level decomposition), \textbf{LineSplit} (newline-based decomposition), and \textbf{BoN} (treating the entire solution as a single step). These are all standard and the most commonly used methods (see Sec.~\ref{sec:prior_decomp}). 

\paragraph{Metrics.} We evaluate two key metrics: \textbf{Pass@k}, the proportion of problems solved within a sample budget $k$, and \textbf{Pass@token}, the proportion solved within a given token budget. Note that $k$ refers to the sample budget, \emph{not thousands of samples}, and error proportion refers to proportion not solved. We use $\alpha_0=0.15, \sigma=1.0$, and temperature $\tau=0.2$ by default for \decomp unless otherwise specified. 

\begin{figure}[ht]
    \centering
    \begin{minipage}{0.31\textwidth}
        \centering
        \includegraphics[width=\linewidth]{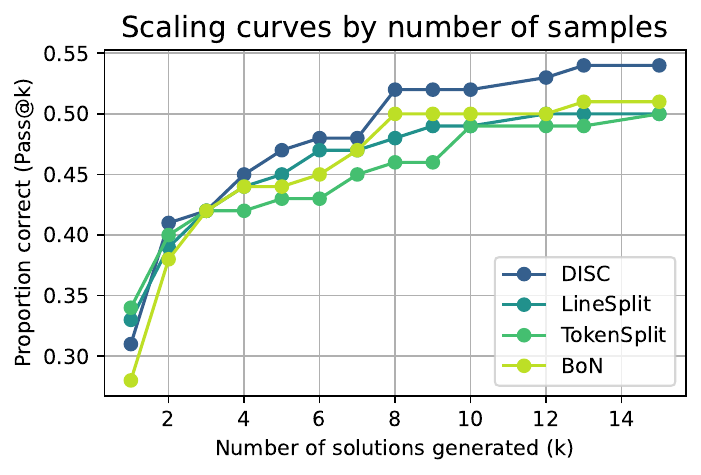}
    \end{minipage}
    \hfill
    \begin{minipage}{0.35\textwidth}
        \centering
        \includegraphics[width=\linewidth]{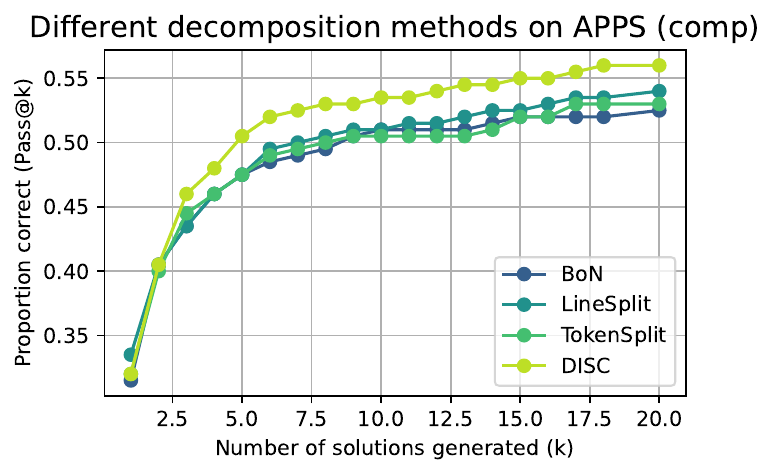}
    \end{minipage}
    \hfill
    \begin{minipage}{0.31\textwidth}
        \centering
        \includegraphics[width=\linewidth]{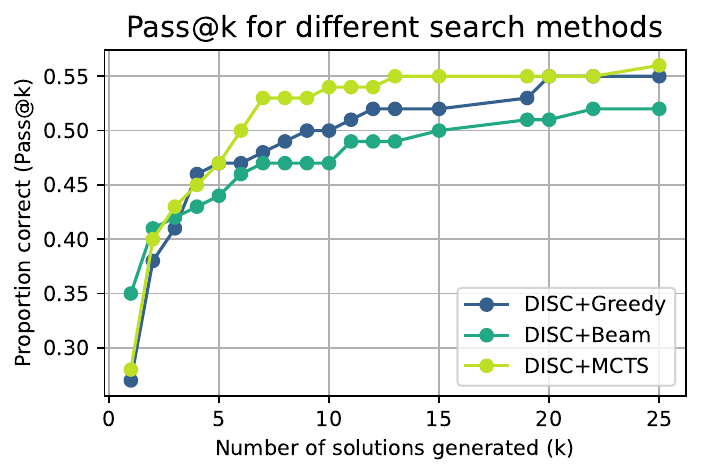}
    \end{minipage}
    \caption{\textbf{Comparison of Pass@k performance on APPS using gpt-4o-mini.} (Left) self generated validation tests, (Middle) with ground truth tests,  (Right) with different search methods.}
    \vspace{-0.1in}
    \label{fig:passk_apps_comp}
\end{figure}

\paragraph{Performance.}  
Across all benchmarks, \decomp consistently delivers stronger performance and better scaling under both fixed token budgets (Fig.~\ref{fig:token_comparison}) and fixed sample budgets (Fig.~\ref{fig:passk_apps_comp}). For example, on APPS, the pass@10 error proportion decreases from 0.50 to 0.475; on MATH500, from 0.15 to 0.14; and on LiveCodeBench, from 0.57 to 0.51. These correspond to a \textbf{5.0\%, 6.7\%, and 10.5\% reduction in error} relative to the best baseline, respectively. 
These improvements are particularly meaningful on more challenging benchmarks, where performance gains are harder to achieve, demonstrating \decomp’s effectiveness in guiding search toward high-reward regions. Extended results and analyses are provided in App.~\ref{sec:apps_extended},~\ref{sec:math500_extended}, and~\ref{sec:livecodebench_extended}.



\begin{figure}[ht]
    \centering
    \begin{minipage}{0.32\textwidth}
        \centering
        \includegraphics[width=\linewidth]{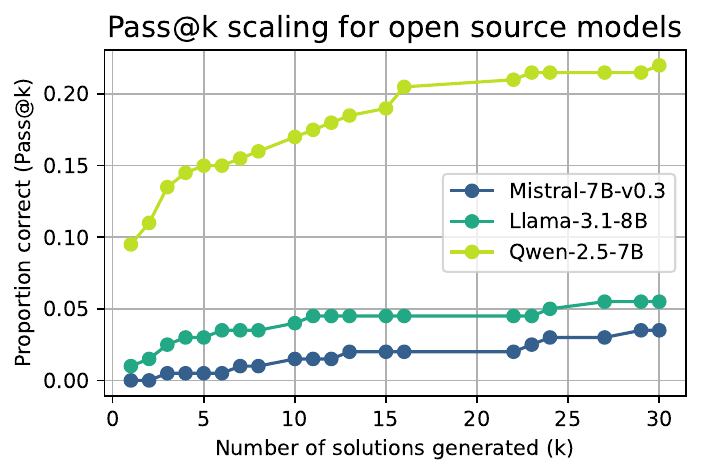}
    \end{minipage}
    \hfill
    \begin{minipage}{0.32\textwidth}
        \centering
        \includegraphics[width=\linewidth]{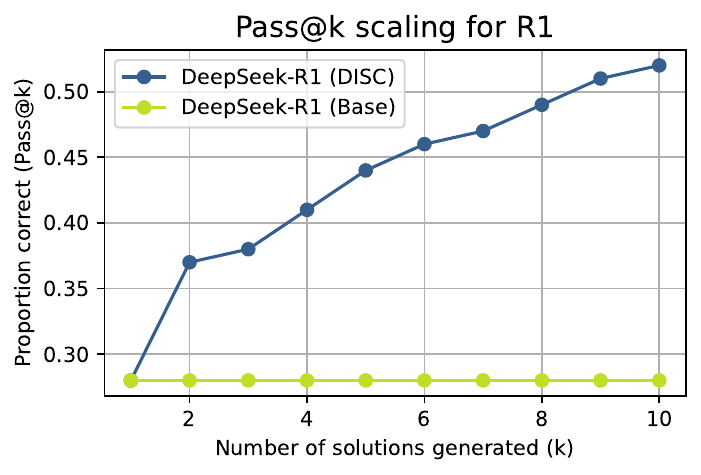}
    \end{minipage}
    \hfill
    \begin{minipage}{0.32\textwidth}
        \centering
        \includegraphics[width=\linewidth]{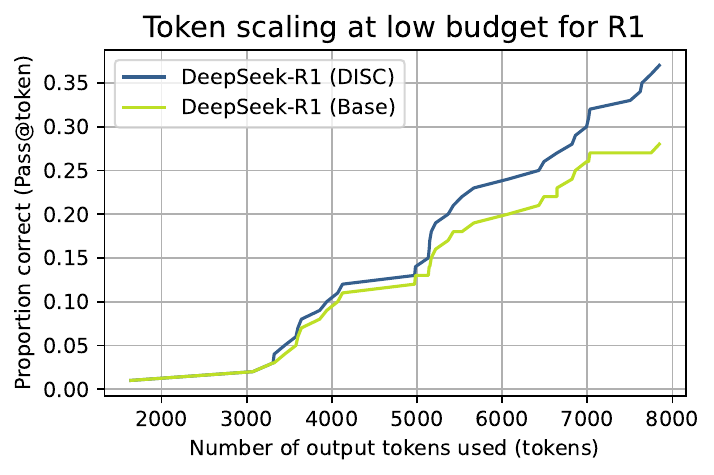}
    \end{minipage}
    \vspace{-0.3cm}
    \caption{\textbf{Inference scaling for different models, including reasoning models, on APPS.} \textbf{(Left} Pass@k for open-source models. Colors correspond to different models, with Pass@1 being the base model performance. \textbf{(Middle)} Pass@k for R1, with green line as base Pass@1 performance. \textbf{(Right)} Pass@token for R1. \decomp almost \textbf{doubles} base performance across all models.}
    \label{fig:open_source_three}
\end{figure}

\paragraph{Model Generality.}  
\decomp also provides substantial gains across a range of models, including open-source LLMs. For instance, it improves LLaMA’s pass@10 rate from 0.01 to 0.04—a \textbf{300\% relative increase}. Similarly, it boosts Mistral’s performance from 0.0 to 0.02, and Qwen’s from 0.095 to 0.17, representing a \textbf{79\% relative increase}. As shown in Fig.~\ref{fig:open_source_three} (left), these improvements hold consistently across sampling budgets, including in low-budget regimes. This demonstrates \decomp’s general applicability and effectiveness even for weaker or resource-constrained models. Additional results and analyses are provided in App.~\ref{sec:model_ablation}.

\paragraph{Reasoning Models.} 
\decomp also yields substantial gains when applied to strong long-form chain-of-thought (CoT) reasoning models such as R1~\citep{guo2025deepseek, chen2025towards}. As shown in Fig.~\ref{fig:open_source_three} (middle), \decomp improves accuracy by over \textbf{85\%} relative to the base R1 model using just 10 samples. Notably, Fig.~\ref{fig:open_source_three} (right) shows that even under a constrained token budget—matched to that of a single sample from the base model—\decomp still achieves over a \textbf{33\% relative improvement}. This demonstrates that \decomp not only scales well with more samples but is also highly effective at identifying and prioritizing critical reasoning steps, leading to stronger performance even in low-resource settings.



\begin{figure}[H]
    \centering
    \includegraphics[width=0.95\textwidth]{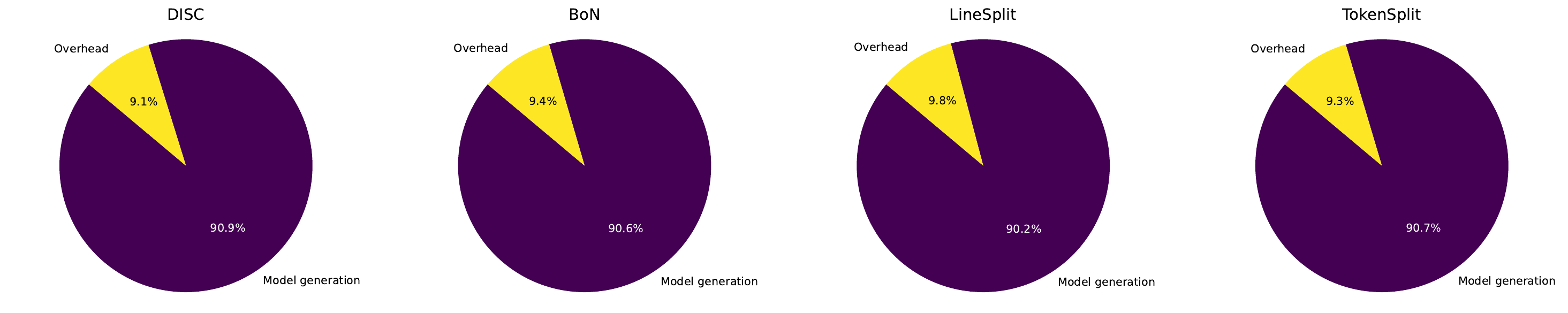}
    \caption{\textbf{Percentage of runtime spent on overhead vs LLM token generation.} \decomp does not increase the runtime overhead significantly.}
    \label{fig:overhead}
\end{figure}

\paragraph{Computational Overhead.} 
\decomp introduces negligible runtime overhead compared to standard decoding baselines, as shown in Fig.~\ref{fig:overhead}. The vast majority of compute time---over 90\% across all settings---is still dominated by LLM token sampling, with only a minor fraction spent on auxiliary operations such as candidate management, z-score normalization, and recursive branching. Despite its dynamic control flow, \decomp maintains a runtime composition nearly identical to that of methods like BoN and LineSplit. Moreover, because \decomp achieves higher success rates with fewer tokens (Fig.~\ref{fig:token_comparison}), its effective runtime per correct solution is even lower. In practice, this makes \decomp both algorithmically efficient and computationally scalable, offering improved search performance without additional inference cost. We provide a detailed runtime discussion in App.~\ref{sec:computational_overhead}.


\paragraph{Self-Generated Validation.}

We further evaluate \decomp in a self-generated validation setting, where ground-truth reward models or curated tests are unavailable~\citep{chen2022codet, chen2023teaching, zhou2023language}. Here, the LLM generates its own unit tests from the problem description, which serve as a \textbf{proxy reward model} for evaluating candidate solutions. This setup offers a \emph{scalable alternative} to costly manual test curation in real-world code generation tasks.

As shown in Fig.~\ref{fig:passk_apps_comp}, \decomp scales effectively under this protocol, achieving a \textbf{54\% relative improvement} over the base model. This demonstrates that \decomp leverages decomposition-based reasoning to produce higher-quality code even when supervision is noisy or incomplete.

However, self-generated validation has limitations: generated tests may be incomplete, inconsistent, or narrow in coverage, potentially biasing performance estimates. While these issues do not affect the observed scaling trends, improving test reliability and robustness remains an important direction. Additional details and results are provided in App.~\ref{sec:val_tests}.

\paragraph{Search.}
\label{sec:search_results}
We demonstrate that search strategies such as \textbf{MCTS} and \textbf{beam search} can be naturally integrated with \decomp using the approach described in Sec.~\ref{sec:disc_for_search}. As shown in Fig.~\ref{fig:search_actualpart}, greedy search tends to explore deeper partitions within the same search budget due to its myopic nature, while MCTS and beam search explore more diverse but shallower paths. Despite similar depth, \textbf{MCTS} outperforms beam search by allocating its search budget more strategically—focusing exploration on more promising candidates—resulting in superior overall performance, as seen in Fig.~\ref{fig:passk_apps_comp}. Furthermore, unlike greedy search, which irrevocably commits to prefixes as they are accepted, MCTS maintains flexibility by exploring committing multiple candidate prefixes in parallel. Additional details and analysis are provided in App.~\ref{sec:search_extended}.

\subsection{Ablation Studies}

\paragraph{Temperature.}


\begin{figure}[ht]
    \centering
    \vspace{-0.3cm}
    \begin{minipage}{0.32\textwidth}
        \centering
        \includegraphics[width=\linewidth]{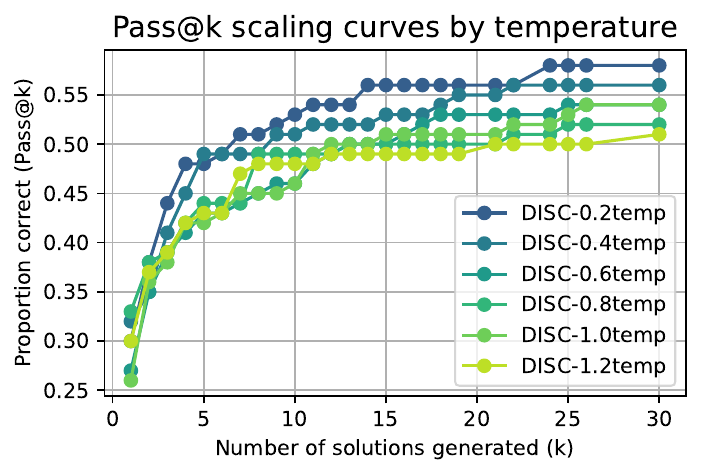}
    \end{minipage}
    \hfill
    \begin{minipage}{0.33\textwidth}
        \centering
        \includegraphics[width=\linewidth]{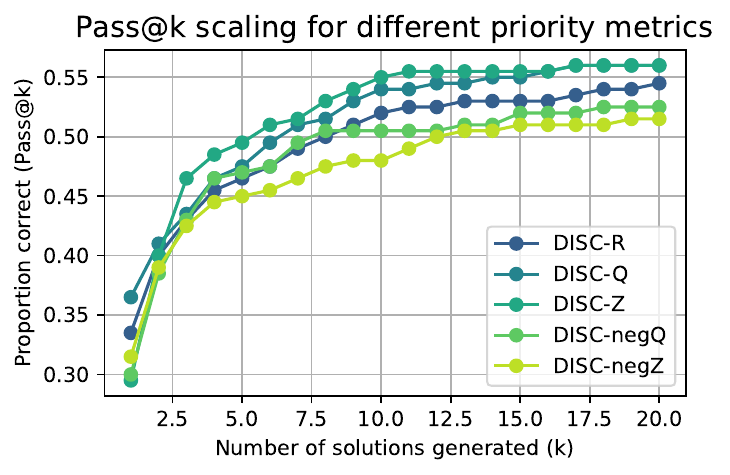}
    \end{minipage}
    \hfill
    \begin{minipage}{0.32\textwidth}
        \centering
        \includegraphics[width=\linewidth]{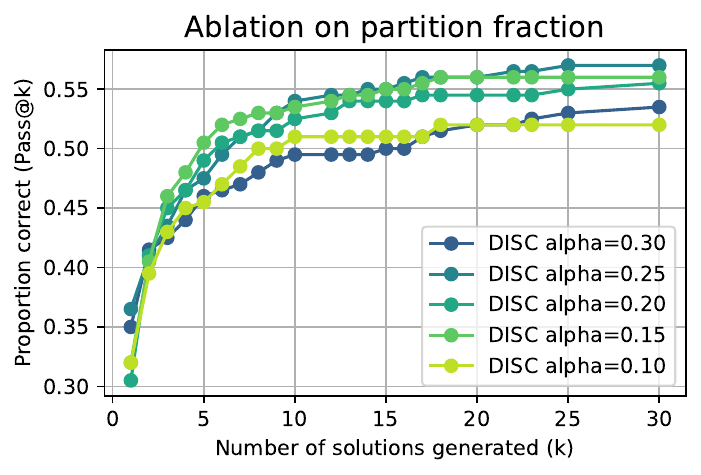}
    \end{minipage}
    \vspace{-0.3cm}
    \caption{\footnotesize{\textbf{Ablation study of \decomp  on APPS with gpt-4o-mini.} (Left) Effect of temperature: unlike BoN and other inference scaling methods, \decomp achieves higher performance at lower temperatures. (Middle) effect of acceptance method (Right). Effect of partition fraction $\alpha_0$: The range $0.15 \leq \alpha_0 \leq 0.25$ appears optimal.}}
    \vspace{-0.3cm}
    \label{fig:decomp_ablation}
\end{figure}
Typically, inference scaling methods achieve optimal performance at temperatures around 0.6–0.8, as increased temperature promotes sample diversity~\citep{wang2024planning}. 
Surprisingly, however, \decomp performs \emph{better at lower temperatures}, as shown in Fig.~\ref{fig:decomp_ablation}. This trend is in stark contrast to BoN (Fig.~\ref{fig:temp_bon_token}), where higher temperatures are generally beneficial.
We believe this phenomenon arises because \decomp depends on  estimating the z-score at each step using sample statistics. Lower temperatures reduce sample variance, leading to more reliable estimates, which in turn improves step selection. 
This is further supported by Fig.~\ref{fig:temp_stdstep}, which shows that lower temperatures yield lower standard deviations per step, indicating increased sampling consistency.
Additional details and analyses can be found in App.~\ref{sec:ab_temp}.


\paragraph{Acceptance Method.}
We perform an ablation study to evaluate whether using the z-score is an effective criterion for accepting candidate prefixes. Specifically, we compare our standard z-score-based acceptance method, \decomp-Z, against four alternative baselines: \decomp-R, which accepts candidates uniformly at random; \decomp-Q, which accepts if the candidate prefix has a lower mean value; \decomp-negQ, which accepts if the candidate has a higher mean; and \decomp-negZ, which accepts if the candidate has a higher z-score (rather than a lower one). As shown in Fig.~\ref{fig:decomp_ablation}, the choice of acceptance criterion substantially influences performance. Among all methods, \decomp-Z achieves the highest performance, while \decomp-negZ performs worse than random selection, underscoring the importance of prioritizing candidates with a higher probability of improvement.
Additional details and analysis are in App.~\ref{sec:ab_prioritymetric}.





\paragraph{Partition Fraction $\alpha_0$.}
As shown in Fig.~\ref{fig:decomp_ablation} and Fig.~\ref{fig:alpha_token}, performance is highest when the partition fraction lies in the range \(0.15 \leq \alpha_0 \leq 0.25\). Smaller values of \(\alpha_0\) generally yield better results because they lead to more conservative proposals—i.e., shorter candidate prefixes. This conservatism is beneficial due to the high cost of prematurely committing to a suboptimal prefix: once a candidate prefix is accepted (in greedy-\decomp), it becomes fixed and cannot be revised. By keeping candidate prefixes short, the algorithm retains more flexibility to correct course in future steps. Additional analysis is provided in App.~\ref{sec:ab_alphafraction}.




\subsection{Analysis and Interpretation}

\begin{figure}[ht]
    \centering
    \vspace{-0.3cm}
    \begin{minipage}{0.32\textwidth}
        \centering
        \includegraphics[width=\linewidth]{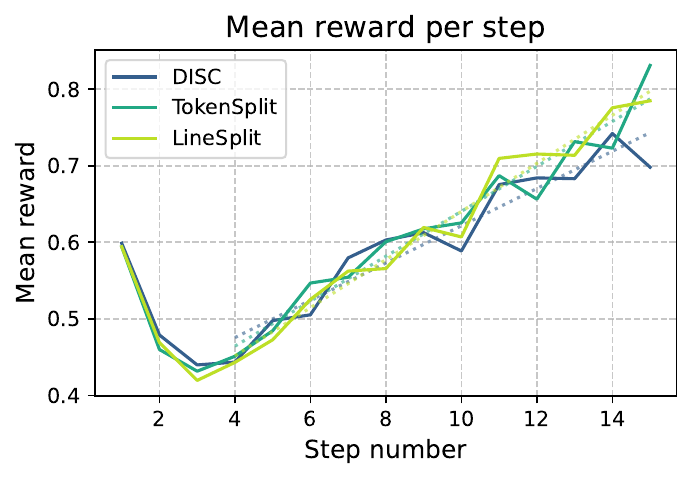}
    \end{minipage}
    \hfill
    \begin{minipage}{0.32\textwidth}
        \centering
        \includegraphics[width=\linewidth]{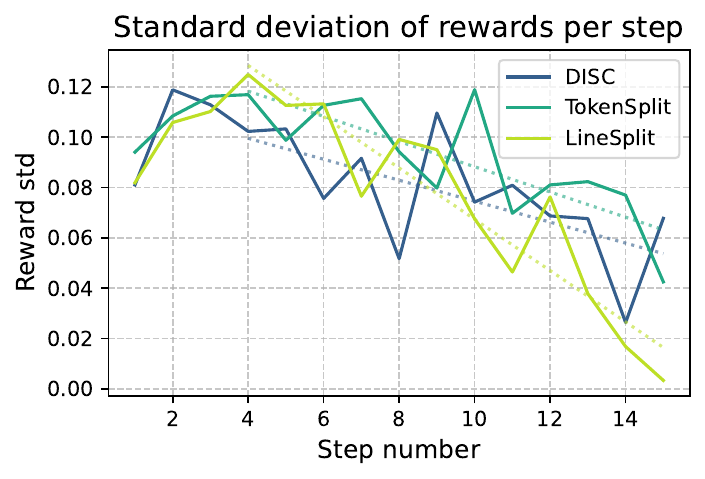}
    \end{minipage}
    \hfill
    \begin{minipage}{0.32\textwidth}
        \centering
        \includegraphics[width=\linewidth]{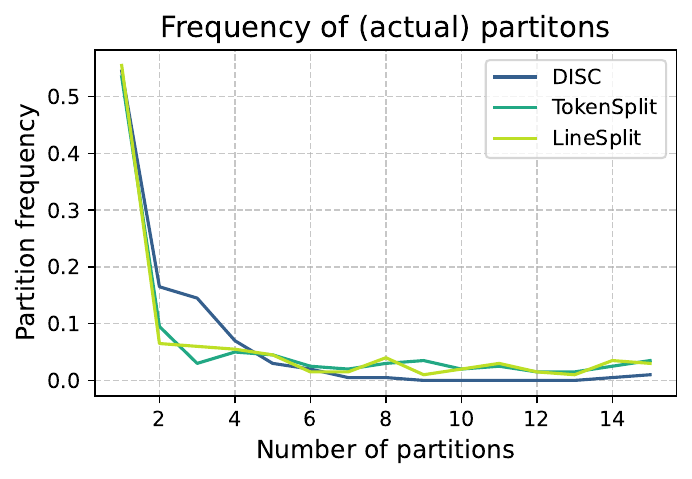}
    \end{minipage}
    \vspace{-0.2cm}
    \caption{\textbf{Analysis of decomposition methods.} Dotted lines fit a linear model to indicate the trend. \textbf{(Left)} Average reward per step: From step 3 onward, higher step counts strongly correlate with increased average reward, demonstrating the effectiveness of decomposition. The dip between steps 1 and 3 likely occurs because simple problems are solved early, preventing further search. \textbf{(Middle)} Standard deviation of rewards per step: Decomposition reduces sampling variance, improving precision at deeper search depths. \textbf{(Right)} Frequency of the number of partition steps that the search algorithm committed to the prefix during the search process.}
    \vspace{-0.3cm}
    \label{fig:decomp_analysis}
\end{figure}

Our results strongly suggest that decomposition—whether line-based, token-based, or \decomp—consistently improves sample quality. Fig.~\ref{fig:decomp_analysis} (left and middle) shows how the mean and variance of sampled rewards evolve with the \textbf{step number}, i.e., the number of steps committed to the prefix during search. As the step number increases, the mean reward improves, indicating that longer committed prefixes lead to higher-quality solutions. At the same time, the variance of rewards decreases, suggesting that committing to a longer prefix also improves the precision of sampling. These trends highlight the benefits of finer-grained decomposition and incremental commitment in guiding the search process more effectively.

Furthermore, \decomp achieves higher performance using fewer committed prefix steps—and thus fewer sampling stages—under a fixed sampling budget. Fig.~\ref{fig:decomp_analysis} (right) shows the distribution of \textbf{actual partitions}, i.e., the number of steps effectively committed under a finite budget. As shown, \decomp typically requires only 1–5 actual partition steps, whereas other methods commit to significantly more. This indicates that \decomp is more efficient at identifying high-impact prefixes, enabling better performance with fewer sampling decisions.

\paragraph{Limitations.}
While DISC demonstrates strong empirical performance, several limitations remain (see Appendix~\ref{app:limitations} for a detailed discussion). In particular, DISC assumes access to a reasonably accurate reward model. For example, in code generation tasks, this requires that either ground-truth validation tests are available or the model can self-generate sufficiently reliable ones. Addressing these limitations—through improved test generation, learned verifiers, and more robust evaluation protocols—presents promising directions for future work.

\begin{tcolorbox}[title=Decomposition and Sample Quality, colframe=low]
\decomp enables more efficient exploration by identifying high-impact prefixes with fewer steps. Incremental prefix commitment not only improves sample quality—yielding higher average rewards—but also reduces reward variance, leading to more stable and reliable outputs under a fixed sampling budget.
\end{tcolorbox}

%% file: sections/related_works.tex
\section{Related Work}
\textbf{Inference scaling.}  
Inference scaling has emerged as a dominant paradigm, driven by the introduction of o1- and r1-like chain-of-thought reasoning models~\citep{snell2024scaling, brown2024large, manvi2024adaptive, leea2025evolving}. Several works examine the trade-off between inference compute and training compute~\citep{guan2025rstarmathsmallllmsmaster, chen2024think}. LLM inference often relies on decomposing complex problems into intermediate reasoning steps, as seen in chain-of-thought (CoT) prompting~\citep{wei2022chain, sprague2024cotcot, wang2024chainofthoughtr} and its variants~\citep{kojima2022large, zhouleast, wangself, li2023making}. We extend inference scaling by introducing a new approach for adaptive compute allocation~\citep{manvi2024adaptive, alabdulmohsin2025recursive, wang2025dynscaling}.

\textbf{LLM reasoning and code generation.}  
LLM reasoning and code generation are central tasks for inference scaling. Evolutionary inference scaling methods have been explored in program generation~\citep{liventsev2023fully, chen2023evoprompting, romera2024mathematical, lehman2023evolution, hemberg2024evolving}. Domain-specific decomposition strategies have been applied in code generation, such as function-based decomposition~\citep{chen2024divide, zenkner2024abstractbeam, levin2025effective}. More broadly, decomposition often involves prompting LLMs to generate subtask completions~\citep{hernandez2recursive, khot2022decomposed, dua2022successive}, which differs from methods that refine a single LLM generation.

\textbf{Reinforcement learning and Monte Carlo methods.}  
Unlike standard RL, our setting resembles a search problem where the goal is to identify the single highest-reward path. Nested Monte Carlo search can accelerate optimal pathfinding ~\citep{cazenave2009nested}. Under the bandit setting, this can be formulated as identifying the arm with the highest \emph{maximum} reward rather than the highest mean reward~\citep{cicirello2005max, carpentier2014extreme}.

%% file: sections/conclusion.tex
\section{Conclusion}  
We introduce \decomp, a dynamic decomposition framework that adaptively partitions solution steps based on first order statistics that capture potential for improvement, improving inference scaling by directing compute toward critical steps while balancing exploration and resource allocation. 
\decomp naturally integrates with search-based methods such as MCTS and beam search, further enhancing performance.
It also identifies challenging steps for LLMs, aiding curriculum learning, fine-tuning, and dataset augmentation. By dynamically adjusting partitioning and step sizes based on available compute, \decomp enables more adaptive and efficient reasoning in large language models, with broad implications for both training and inference optimization.


%% file: sections/appendix.tex

\input{sections/appendix/pythoncode}

\input{sections/appendix/pseudocode}

\input{sections/appendix/ablation_extended}

\input{sections/appendix/experiments_extended}

\input{sections/appendix/limitations}

\input{sections/appendix/search}

\input{sections/appendix/analysis}

\newpage
\section{Compute Resources Used}
All evaluations involving OpenAI proprietary models (e.g., GPT-4o-mini, GPT-3.5) were performed via the OpenAI API. These API calls were made from standard desktop machines using CPU-only inference and incur no dependency on local GPU availability, making the method broadly accessible for replication.

For open-source model experiments (e.g., LLaMA-3.1-8B-Instruct, Mistral, Qwen), inference was conducted on a single NVIDIA A100 GPU. All such experiments were executed sequentially on this GPU, which has 80GB of memory, ensuring consistent and reproducible runtime characteristics across model families.

This setup reflects the lightweight computational overhead of our method and demonstrates that DISC can scale effectively even in constrained or CPU-only environments when using hosted APIs.


\section{Impact Statement}
\paragraph{Positive Societal Impacts}
This work introduces a general and lightweight method—Dynamic Decomposition (DISC)—that significantly improves inference efficiency for large language models (LLMs) without requiring additional training, domain-specific engineering, or specialized hardware. By enabling better performance using fewer samples and tokens, DISC reduces the cost and environmental footprint associated with LLM deployment, making advanced reasoning models more accessible to research groups and developers with limited compute budgets.

Additionally, DISC's ability to prioritize critical reasoning steps has the potential to improve the transparency and interpretability of LLM outputs, which is beneficial in high-stakes applications such as education, scientific reasoning, and assistive tools for programming or mathematics. Its plug-and-play compatibility with open-source models also democratizes access to high-performance inference techniques.

\paragraph{Negative Societal Impacts}
As with all improvements in LLM inference capabilities, this work may accelerate the deployment of models in settings where societal risks are not fully mitigated—such as the automated generation of persuasive misinformation, cheating in educational contexts, or manipulation via high-fidelity language generation. Furthermore, by making LLM inference more efficient, DISC could contribute to increased usage of models without corresponding increases in ethical oversight or alignment safeguards.

Careful integration of this method should therefore include responsible use policies, limitations on deployment domains, and alignment with values such as transparency, fairness, and accountability.

%% file: sections/appendix/pythoncode.tex
\section{Code implementation of \decomp}
\label{app:code_impl}

To aid reproducibility and practical adoption, we provide Python-style pseudocode in this section that closely mirrors our actual implementation. The pseudocode abstracts away low-level engineering details while preserving the core logic, function names, and control flow used in our codebase. This alignment ensures that readers can easily translate the pseudocode into a working implementation or modify it for their own use cases. All key components of the DISC algorithm—including dynamic step refinement, z-score-based acceptance, and integration with search strategies—are reflected in this pseudocode with minimal deviation from the actual code.

\begin{tcolorbox}[breakable, colback=white, title=Python implementation of \decomp]
    \begin{lstlisting}
def dynamic_decomposition(problem, model, reward_model, split_str, complete_solution, fraction, solution_budget, split_metric, stop_threshold=-float("inf"), stop_sum_score=1.0, stop_if_solved=False, ):
    """
    Decomposes the solution using a dynamic binary search approach

    Args:
        problem (Problem): The problem to solve
        model (Model): The model to use for generation
        reward_model (function): The reward model to use for scoring
        split_str (function): The function to use for splitting a string
        complete_solution (function): The function to use for completing a solution
        fraction (float): The fraction to split the string
        solution_budget (int): The maximum number of solutions to generate
        split_metric (function): The metric to use for splitting
        stop_threshold (float): The threshold to stop splitting
        stop_sum_score (float): The sum score to stop generating completions
        stop_if_solved (bool): Whether to stop if the problem is solved
    """

    # Initialize results and decomposition steps
    decomp_return = {
        "generated_solutions": [],
        "decomposition": []
    }

    while len(decomp_return["generated_solutions"]) < solution_budget:
        # Combine all previous steps into an intermediate solution
        intermediate_solution = "".join([step["step_str"] for step in decomp_return["decomposition"]])
        new_scores = []
        best_solution = None
        best_completion = None
        best_score = -float("inf")
        sum_score = 0.0

        # 1) Generate completions until we generate enough samples to estimate the split metric
        while sum_score < stop_sum_score:
            proposed_completion = complete_solution(problem, intermediate_solution, model)
            proposed_solution = intermediate_solution + proposed_completion
            decomp_return["generated_solutions"].append(proposed_solution)

            # Update scores
            proposed_score = reward_model(proposed_solution)
            new_scores.append(proposed_score)
            sum_score += proposed_score

            # Track the best solution
            if proposed_score > best_score:
                best_solution = proposed_solution
                best_score = proposed_score
                best_completion = proposed_completion

            # Stop early if problem is solved
            if stop_if_solved and proposed_score >= 1.0:
                decomp_return["decomposition"].append({"step_str": proposed_completion})
                return decomp_return

        new_metric = split_metric(new_scores)
        last_metric = decomp_return["decomposition"][-1]["metric"] if decomp_return["decomposition"] else None

        # Determine the split target. We always split the step with the highest metric
        is_split_new_step = last_metric is None or new_metric >= last_metric
        split_target = decomp_return["decomposition"][-1]["step_str"] if not is_split_new_step else best_completion

        # 3) Attempt to split the target
        split_result = split_str(split_target, fraction)
        if not split_result: # If we can't split the target, we're done
            decomp_return["decomposition"].append({"step_str": best_completion, "metric": new_metric})
            return decomp_return

        # Update decomposition based on split
        part1, part2 = split_result
        if is_split_new_step:
            decomp_return["decomposition"].append({"step_str": part1, "metric": new_metric})
            # Stopping condition based on threshold
            if new_metric < stop_threshold: 
                decomp_return["decomposition"].append({"step_str": part2})
                return decomp_return
        else:
            decomp_return["decomposition"][-1] = {"step_str": part1, "metric": last_metric}

    return decomp_return
\end{lstlisting}
\end{tcolorbox}

%% file: sections/appendix/pseudocode.tex
\section{Pseudocode for \decomp}
\begin{algorithm}[h!]
\caption{Dynamic Decomposition}
\label{alg:dynamic_decomposition}
\begin{algorithmic}[1]
    \State {\bfseries Input:} Problem instance $\textcolor{algcolor3}{x}$, reward model $\textcolor{algcolor4}{r}$, 
    partition function $\textcolor{algcolor6}{f}$, LLM policy model $\textcolor{algcolor6}{\pi}$,  
    partition fraction $\textcolor{algcolor6}{\alpha}$, solution budget $\textcolor{algcolor6}{B}$, priority metric $\textcolor{algcolor5}{h}$, 
    metric stopping precision $\textcolor{algcolor5}{\theta}$, sampling stopping threshold $\textcolor{algcolor4}{\sigma}$, 
    is inference mode \textbf{b}\textsubscript{\text{inference}}
    \State {\bfseries Output:} Final decomposition $\textcolor{algcolor6}{D}$

    \State Initialize $\textcolor{algcolor6}{D} \gets \{\text{generated\_solutions}: \varnothing, \text{decomposition}: \varnothing\}$
    \State \textcolor{algcolor1}{\# Decompose the solution recursively until we reach the desired precision $\textcolor{algcolor5}{\theta}$ or run out of budget $\textcolor{algcolor6}{B}$}
    \While{$| \textcolor{algcolor6}{D}.\text{generated\_solutions} | < \textcolor{algcolor6}{B}$}
        \State $\textcolor{algcolor3}{y_{\text{intermediate}}} \gets \text{Concatenate}([ \text{step.step\_str} \ \forall \ \text{step} \in \textcolor{algcolor6}{D}.\text{decomposition} ])$ 
        \State \hspace*{\fill} \textcolor{algcolor2}{$\vartriangleright$ Concatenate previous steps to form intermediate solution}
        \State $\textcolor{algcolor4}{R_{\text{new}}} \gets \varnothing$ \hspace*{\fill} \textcolor{algcolor2}{$\vartriangleright$ Record rewards of completions}
        \State $\textcolor{algcolor3}{\text{best}.y_{\text{final}}} \gets \text{None}$, $\textcolor{algcolor3}{\text{best}.y_{\text{completion}}} \gets \text{None}$,  $\textcolor{algcolor4}{\text{best}.r} \gets -\infty$ 
        \hspace*{\fill} \textcolor{algcolor2}{\hspace*{\fill} $\vartriangleright$ Track the best completion}

        \State \textcolor{algcolor1}{\# Step 1: Generate completions until we have enough samples to estimate the splitting metric. Here we use a geometric sampling distribution}
        \While{$\operatorname{sum}( \textcolor{algcolor4}{R_{\text{new}}} ) < \textcolor{algcolor4}{\sigma}$}
            \State $\textcolor{algcolor3}{y_{\text{completion}}} \gets \textcolor{algcolor6}{\pi}(\cdot | \textcolor{algcolor3}{x}, \textcolor{algcolor3}{y_{\text{intermediate}}})$
            \State $\textcolor{algcolor3}{y_{\text{proposed}}} \gets \textcolor{algcolor3}{y_{\text{intermediate}}} \oplus \textcolor{algcolor3}{y_{\text{completion}}}$
            \State Append $\textcolor{algcolor3}{y_{\text{proposed}}}$ to $\textcolor{algcolor6}{D}.\text{generated\_solutions}$
            \State $\textcolor{algcolor4}{r_{\text{proposed}}} \gets \textcolor{algcolor4}{r}(\textcolor{algcolor3}{y_{\text{proposed}}})$
            \State Append $\textcolor{algcolor4}{r_{\text{proposed}}}$ to $\textcolor{algcolor4}{R_{\text{new}}}$
            \If{$\textcolor{algcolor4}{r_{\text{proposed}}} > \textcolor{algcolor4}{\text{best}.r}$}
                \State $\textcolor{algcolor3}{\text{best}.y_{\text{final}}} \gets \textcolor{algcolor3}{y_{\text{proposed}}}$, $\textcolor{algcolor3}{\text{best}.y_{\text{completion}}} \gets \textcolor{algcolor3}{y_{\text{completion}}}$
                \State $\textcolor{algcolor4}{\text{best}.r} \gets \textcolor{algcolor4}{r_{\text{proposed}}}$
            \EndIf
            \If{$\textbf{b}_{\text{inference}} \ \textbf{and} \ \textcolor{algcolor4}{r_{\text{proposed}}} = 1.0$} 
                \State Append $\{\text{step\_str}: \textcolor{algcolor3}{y_{\text{completion}}} \}$ to $\textcolor{algcolor6}{D}.\text{decomposition}$
                \State \textbf{Return} $\textcolor{algcolor6}{D}$ \hspace*{\fill} \textcolor{algcolor2}{$\vartriangleright$ Exit if problem is solved}
            \EndIf
        \EndWhile

        \State \textcolor{algcolor1}{\# Step 2: Compute splitting metric}
        \State $\textcolor{algcolor5}{z_{\text{new}}} \gets \textcolor{algcolor5}{h}(\textcolor{algcolor4}{R_{\text{new}}})$
        \State $\textcolor{algcolor5}{z_{\text{last}}} \gets \textcolor{algcolor6}{D}.\text{decomposition}[-1].\textcolor{algcolor5}{z} \ \textbf{if} \ \textcolor{algcolor6}{D}.\text{decomposition} \neq \varnothing \ \textbf{else} \ -\infty$

        \State \textcolor{algcolor1}{\# Step 3: Split the step with the higher metric}
        \State \textbf{b}\textsubscript{\text{split new step}} $\gets$ $\textcolor{algcolor5}{z_{\text{new}}} \geq \textcolor{algcolor5}{z_{\text{last}}}$
        \State $\textcolor{algcolor3}{y_{\text{target step}}} \gets \textcolor{algcolor3}{\text{best}.y_{\text{completion}}} \ \textbf{if} \ \textbf{b}\textsubscript{\text{split new step}} \ \textbf{else} \ \textcolor{algcolor6}{D}.\text{decomposition}[-1].\text{step\_str}$
        
        \State $\textcolor{algcolor3}{y_1}, \textcolor{algcolor3}{y_2} \gets \textcolor{algcolor6}{f}(\textcolor{algcolor3}{y_{\text{target step}}}, \textcolor{algcolor6}{\alpha})$
        \If{$\textcolor{algcolor3}{y_1} = \text{None} \ \textbf{or} \ \textcolor{algcolor3}{y_2} = \text{None}$}
            \State Append $\{\text{step\_str}: \textcolor{algcolor3}{y_{\text{completion}}}, \text{metric}: \textcolor{algcolor5}{z_{\text{new}}} \}$ to $\textcolor{algcolor6}{D}.\text{decomposition}$
            \State \textbf{Return} $\textcolor{algcolor6}{D}$ \hspace*{\fill} \textcolor{algcolor2}{$\vartriangleright$ Exit if we cannot do a finer split}
        \EndIf
        \If{$\textbf{b}_{\text{split new step}}$} 
            \State Append $\{\text{step\_str}: \textcolor{algcolor3}{y_1}, \text{metric}: \textcolor{algcolor5}{z_{\text{new}}} \}$ to $\textcolor{algcolor6}{D}.\text{decomposition}$
            \hspace*{\fill} \textcolor{algcolor2}{$\vartriangleright$ Add new step}
            \If{$\textcolor{algcolor5}{z_{\text{new}}} < \textcolor{algcolor5}{\theta}$} 
                \State Append $\{\text{step\_str}: \textcolor{algcolor3}{y_2} \}$ to $\textcolor{algcolor6}{D}.\text{decomposition}$
                \State \textbf{Return} $\textcolor{algcolor6}{D}$ \hspace*{\fill} \textcolor{algcolor2}{$\vartriangleright$ Exit if all metrics are smaller than precision}
            \EndIf
        \Else 
            \State $\textcolor{algcolor6}{D}.\text{decomposition}[-1] \gets \{\text{step\_str}: \textcolor{algcolor3}{y_1}, \text{metric}: \textcolor{algcolor5}{z_{\text{last}}} \}$
            \hspace*{\fill} \textcolor{algcolor2}{$\vartriangleright$ Split last step}
        \EndIf
    \EndWhile

    \State \textbf{Return} $\textcolor{algcolor6}{D}$
\end{algorithmic}
\end{algorithm}

\newpage

%% file: sections/appendix/ablation_extended.tex
\section{Ablation studies}

\subsection{Ablation on Temperature}
\label{sec:ab_temp}

We conduct an ablation study to analyze the effects of temperature on \decomp and BoN. Temperature controls the randomness of token sampling in autoregressive models, influencing both exploration and consistency. Higher temperatures encourage more diverse outputs, whereas lower temperatures yield more deterministic generations. To examine its impact, we evaluate \decomp and BoN on a 100-problem subset of APPS (the first 100 problems) using gpt-4o-mini.

\begin{figure}[H]
    \centering

    \begin{minipage}[t]{0.48\linewidth}
        \centering
        \includegraphics[width=\linewidth]{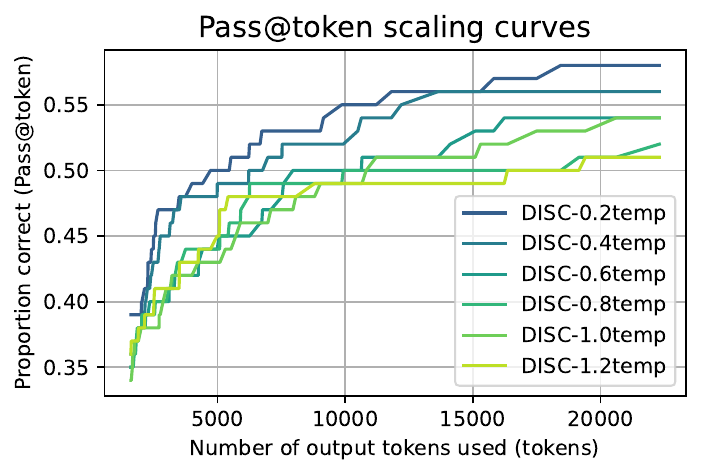}
        \caption{\textbf{Pass@token scaling curve for different temperatures on APPS using gpt-4o-mini}. The lower the temperature, the stronger the \decomp performance.}
        \label{fig:temp_token}
    \end{minipage}%
    \hfill
    \begin{minipage}[t]{0.48\linewidth}
        \centering
        \includegraphics[width=\linewidth]{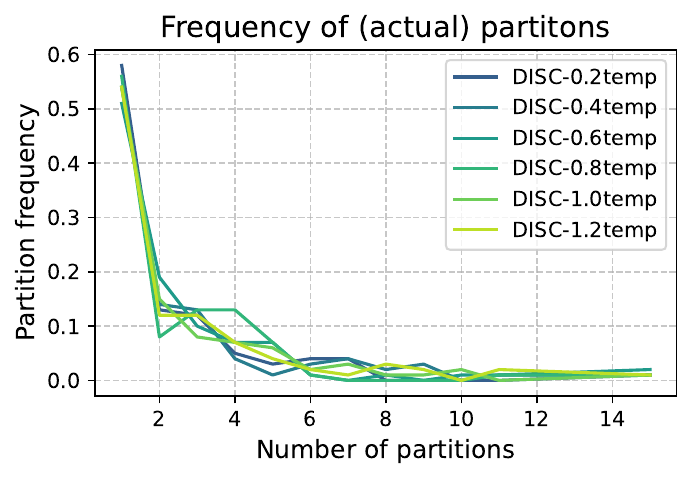}
        \caption{\textbf{Partition frequency of \decomp with different temperatures on APPS using gpt-4o-mini}.}
        \label{fig:temp_actualpart}
    \end{minipage}
    
    \vspace{0.5cm}
    
    \begin{minipage}[t]{0.48\linewidth}
        \centering
        \includegraphics[width=\linewidth]{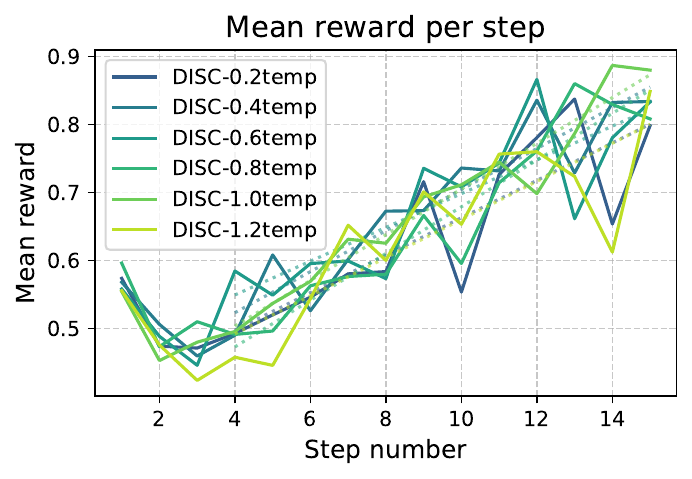}
        \caption{\textbf{Mean reward per step of \decomp with different temperatures on APPS using gpt-4o-mini}. The mean reward scales linearly with step number.}
        \label{fig:temp_rewardstep}
    \end{minipage}%
    \hfill
    \begin{minipage}[t]{0.48\linewidth}
        \centering
        \includegraphics[width=\linewidth]{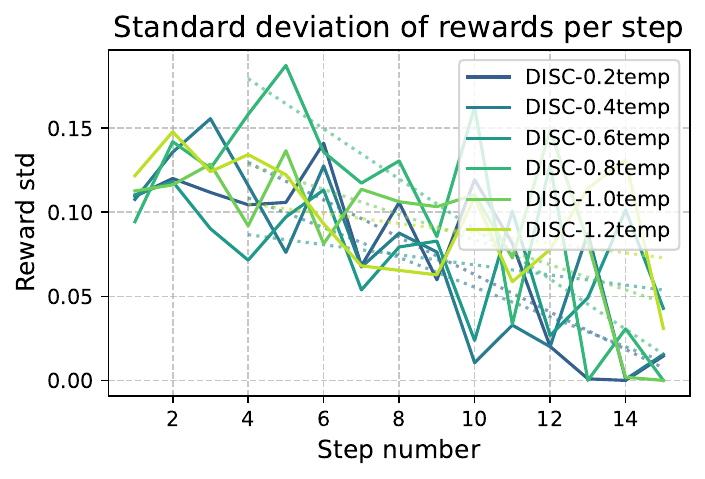}
        \caption{\textbf{Mean standard deviation per step of \decomp with different temperatures on APPS using gpt-4o-mini}. Lower temperature means lower average standard deviation.}
        \label{fig:temp_stdstep}
    \end{minipage}
    
    \vspace{0.5cm}
    
    \begin{minipage}[t]{0.48\linewidth}
        \centering
        \includegraphics[width=\linewidth]{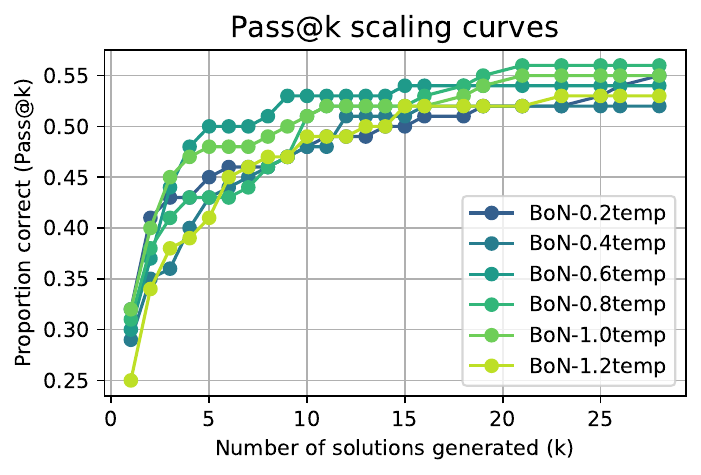}
        \caption{\textbf{Pass@k scaling curve for different temperatures on APPS using gpt-4o-mini for BoN}. A temperature around 0.6–0.8 leads to the best performance and balance between diversity and consistency.}
        \label{fig:temp_bon_passk}
    \end{minipage}%
    \hfill
    \begin{minipage}[t]{0.48\linewidth}
        \centering
        \includegraphics[width=\linewidth]{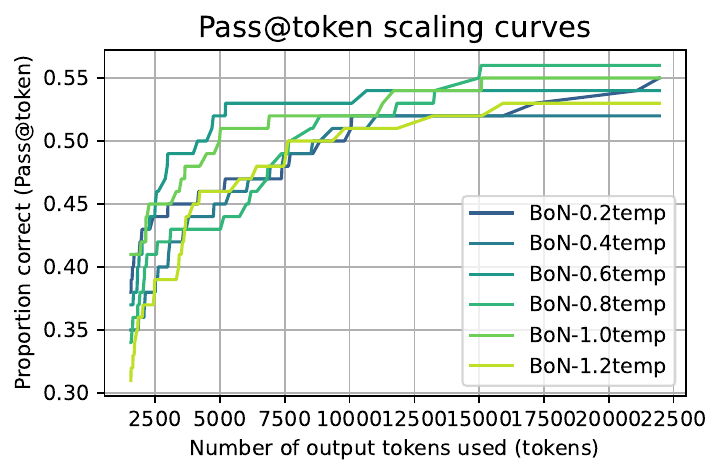}
        \caption{\textbf{Pass@token scaling curve for different temperatures on APPS using gpt-4o-mini for BoN}. A temperature around 0.6–0.8 leads to the best performance and balance between diversity and consistency.}
        \label{fig:temp_bon_token}
    \end{minipage}

\end{figure}

\Cref{fig:temp_token} presents the Pass@token scaling curve for \decomp across different temperatures. The results indicate that lower temperatures lead to improved performance, as \decomp benefits from more deterministic step selection. Unlike BoN, which relies on broad solution sampling, \decomp dynamically refines steps, making stable token probabilities advantageous.

\Cref{fig:temp_actualpart} illustrates the frequency of actual partitions made by \decomp at different temperatures. As temperature increases, the number of partitions fluctuates more, suggesting that high temperature introduces instability in step selection. Lower temperatures provide more structured decomposition, reducing unnecessary subdivisions.

In \Cref{fig:temp_rewardstep}, we visualize the mean reward per step. The trend shows a linear increase in reward as step number grows, demonstrating that deeper decomposition results in progressively better solutions. This reinforces that \decomp effectively allocates computation towards refining difficult steps.

The mean standard deviation per step is shown in \Cref{fig:temp_stdstep}. Lower temperatures yield lower standard deviations, confirming that \decomp benefits from reduced variability in sample quality. This consistency allows for more reliable prioritization of difficult steps, enhancing overall inference efficiency.

For comparison, \Cref{fig:temp_bon_token} and \Cref{fig:temp_bon_passk} display Pass@token and Pass@k scaling curves for BoN across different temperatures. Unlike \decomp, BoN achieves peak performance at a temperature around 0.6-0.8, balancing diversity and consistency. Higher temperatures increase exploration but degrade precision, while lower temperatures hinder sample diversity, reducing the probability of obtaining high-quality completions.

These findings highlight the fundamental difference between \decomp and BoN: \decomp benefits from lower variance and stable decomposition, while BoN relies on broader exploration facilitated by moderate temperature settings. As a result, optimal temperature settings differ significantly between these methods, with \decomp favoring deterministic sampling and BoN requiring a balance between diversity and coherence.

\subsection{Acceptance Method}
\label{sec:ab_prioritymetric}

\begin{figure}[H]
    \centering

    \begin{minipage}[t]{0.48\linewidth}
        \centering
        \includegraphics[width=\linewidth]{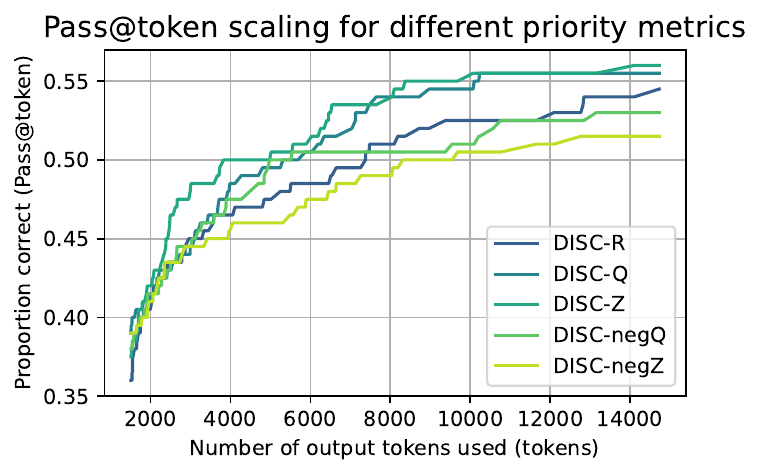}
        \caption{\textbf{Token level comparison of different priority metrics on \decomp in the APPS setting with gpt-4o-mini.} Both Q and Z based priority metrics perform well.}
        \label{fig:metric_token}
    \end{minipage}%
    \hfill
    \begin{minipage}[t]{0.48\linewidth}
        \centering
        \includegraphics[width=\linewidth]{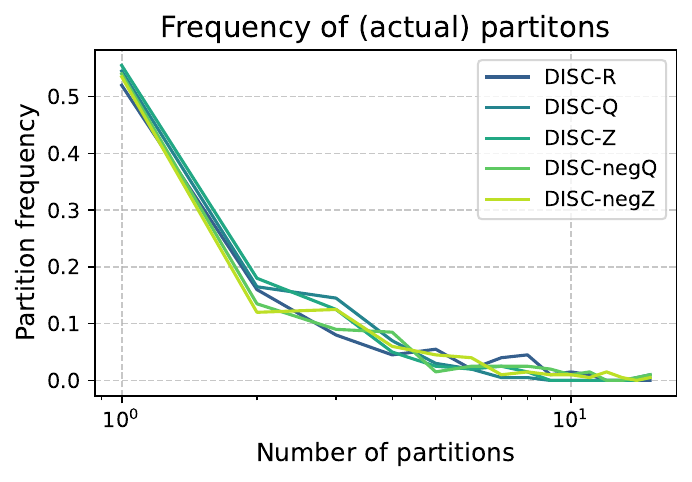}
        \caption{\textbf{Partition frequency of \decomp with different priority metrics on APPS using gpt-4o-mini}.}
        \label{fig:metric_actualpart}
    \end{minipage}

    \vspace{0.5cm}

    \begin{minipage}[t]{0.48\linewidth}
        \centering
        \includegraphics[width=\linewidth]{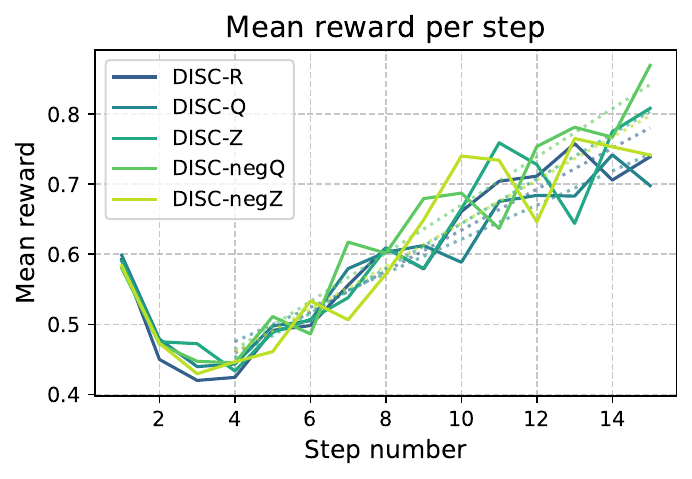}
        \caption{\textbf{Mean reward per step of \decomp with different priority metrics on APPS using gpt-4o-mini}. All metrics display strong correlation between step depth and the mean reward.}
        \label{fig:metric_rewardstep}
    \end{minipage}%
    \hfill
    \begin{minipage}[t]{0.48\linewidth}
        \centering
        \includegraphics[width=\linewidth]{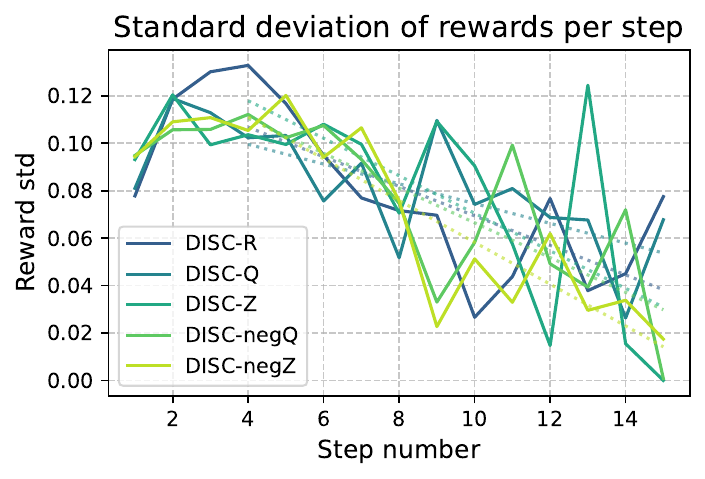}
        \caption{\textbf{Mean standard deviation per step of \decomp with different priority metrics on APPS using gpt-4o-mini}. All metrics display correlation between step depth and the standard deviation.}
        \label{fig:metric_stdstep}
    \end{minipage}

\end{figure}

We analyze the impact of different \textbf{acceptance methods} on \decomp's performance. These methods govern whether a candidate prefix is accepted for further decomposition, directly influencing efficiency, stability, and final solution quality.

We begin by introducing notation to characterize the distribution of rewards obtained when sampling completions from a given prefix $\boldsymbol{p}$. Let $R_{\boldsymbol{p}}$ denote the random variable of the reward of a completion sampled from the LLM policy $\pi$ conditioned on prefix $\boldsymbol{p}$:
\[
R_{\boldsymbol{p}} := R(\boldsymbol{p} \cdot \boldsymbol{s}), \quad \text{where } \boldsymbol{s} \sim \pi(\cdot \mid \boldsymbol{p}).
\]
Let $F_{\boldsymbol{p}}$, $\mu_{\boldsymbol{p}}$, and $\sigma_{\boldsymbol{p}}$ be the cumulative distribution function (CDF), mean, and standard deviation of $R_{\boldsymbol{p}}$, respectively.

To estimate these quantities in practice, we sample $M$ completions from the policy:
\[
Y_{\boldsymbol{p}} = \left\{ \left(\boldsymbol{p} \cdot \boldsymbol{s}^i, \, R(\boldsymbol{s}^i) \right) \;\middle|\; \boldsymbol{s}^i \sim \pi(\cdot \mid \boldsymbol{p}) \right\}_{i=1}^{M}.
\]
From these samples, we compute the empirical mean and standard deviation:
\[
\mu_{\boldsymbol{p}} = \frac{1}{M} \sum_{i=1}^{M} R(\boldsymbol{s}^i), \quad 
\sigma_{\boldsymbol{p}} = \sqrt{ \frac{1}{M} \sum_{i=1}^{M} \left( R(\boldsymbol{s}^i) - \mu_{\boldsymbol{p}} \right)^2 }.
\]
Let \( r^{(1)}_{\boldsymbol{p}} = \max_{i} R(\boldsymbol{s}^i) \) denote the sample maximum reward observed for prefix \( \boldsymbol{p} \).

We evaluate \decomp on the first 200 competition-level APPS problems using \texttt{gpt-4o-mini} at a fixed temperature of \textbf{0.8}. The following acceptance methods are compared:
\begin{itemize}
    \item \textbf{\decomp-Z}: accepts if the candidate prefix $\boldsymbol{c}$ has a lower z-score than the base prefix $\boldsymbol{b}$.
    \item \textbf{\decomp-Q}: accepts if the candidate prefix has a lower mean reward, i.e., $\mu_{\boldsymbol{c}} < \mu_{\boldsymbol{b}}$.
    \item \textbf{\decomp-negZ}: accepts if the candidate has a higher z-score.
    \item \textbf{\decomp-negQ}: accepts if the candidate has a higher mean reward.
    \item \textbf{\decomp-R}: accepts a candidate uniformly at random.
\end{itemize}

\paragraph{Z-score Based Acceptance (\decomp-Z).}
For a base prefix $\boldsymbol{b}$ and candidate prefix $\boldsymbol{c}$, we estimate the z-score of the sample maximum \( r^{(1)}_{\boldsymbol{c}} \) as:
\[
z_{\boldsymbol{c}} = \frac{r^{(1)}_{\boldsymbol{c}} - \mu_{\boldsymbol{c}}}{\sigma_{\boldsymbol{c}}}, \quad \text{so } \mathbb{P}[R_{\boldsymbol{c}} > r^{(1)}_{\boldsymbol{c}}] = 1 - F_{\boldsymbol{c}}(r^{(1)}_{\boldsymbol{c}}).
\]
We accept $\boldsymbol{c}$ if $z_{\boldsymbol{c}} < z_{\boldsymbol{b}}$. A lower z-score implies a greater tail probability mass and thus a higher chance of improvement.

\paragraph{Q-based Acceptance (\decomp-Q).}
This method accepts $\boldsymbol{c}$ if its sample mean is lower than that of $\boldsymbol{b}$:
\[
\mu_{\boldsymbol{c}} < \mu_{\boldsymbol{b}}.
\]
While simple, this method compares absolute expected values without accounting for reward variance, making it less robust to noise in $\pi$’s samples.

\paragraph{Empirical Comparison.}
\Cref{fig:metric_token} shows token-level performance across methods. \decomp-Z significantly outperforms alternatives, highlighting the value of normalizing reward advantage by variance.

\Cref{fig:metric_actualpart} shows that \decomp-Z produces fewer but more meaningful partitions, suggesting better allocation of compute.

\Cref{fig:metric_rewardstep} shows that \decomp-Z achieves sharper reward gains early on, indicating better prioritization of impactful refinements.

\Cref{fig:metric_stdstep} reports step-wise reward variance. \decomp-Z yields the most stable and consistent improvements.

\begin{tcolorbox}[title=Why Z-score Works Best, colframe=low]
Z-score based acceptance balances the mean and variance of reward samples, estimating the probability of improvement in a statistically grounded way. This leads to more reliable and efficient decomposition decisions.
\end{tcolorbox}

Overall, our results confirm that \textbf{acceptance method design is critical for dynamic decomposition}. Among all tested methods, \decomp-Z consistently delivers the best performance and compute efficiency.

\subsection{Model Ablation}
\label{sec:model_ablation}

\begin{figure}[H]
    \centering

    \begin{minipage}[t]{0.48\linewidth}
        \centering
        \includegraphics[width=\linewidth]{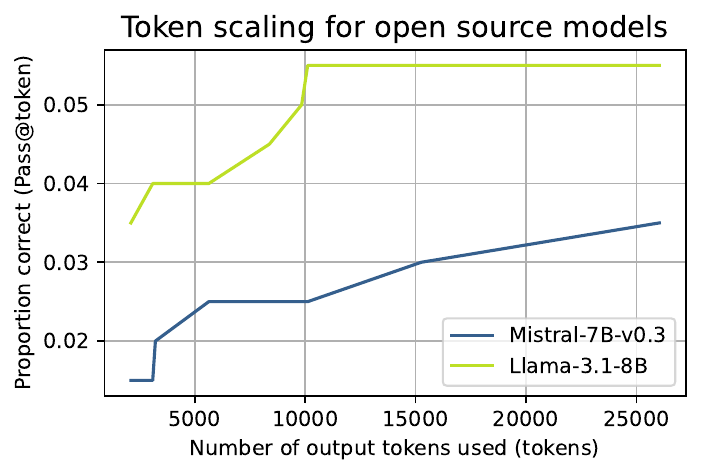}
        \caption{\textbf{Pass@token scaling curve for open source models with \decomp on APPS}. \decomp also demonstrates strong performance gains with open source models.}
        \label{fig:opensource_token}
    \end{minipage}%
    \hfill
    \begin{minipage}[t]{0.48\linewidth}
        \centering
        \includegraphics[width=\linewidth]{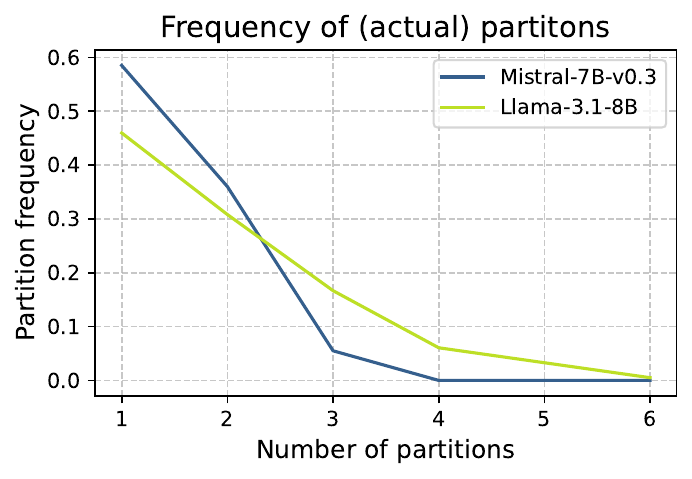}
        \caption{\textbf{Partition frequency of \decomp with open source models on APPS}.}
        \label{fig:open_actualpart}
    \end{minipage}

    \vspace{0.5cm}

    \begin{minipage}[t]{0.48\linewidth}
        \centering
        \includegraphics[width=\linewidth]{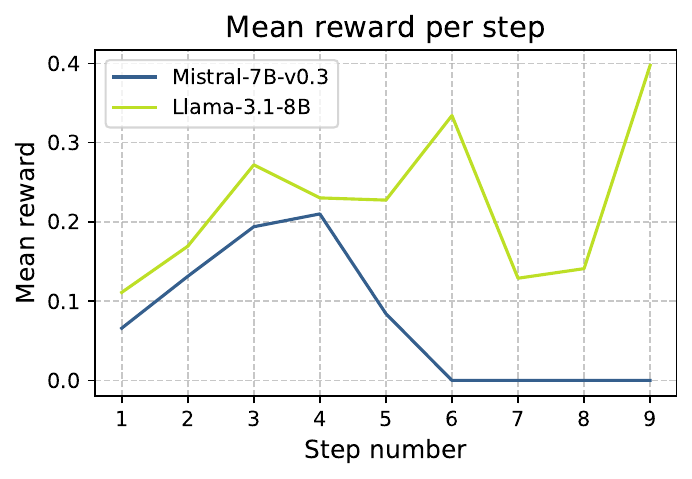}
        \caption{\textbf{Mean reward per step of \decomp with open source models on APPS}.}
        \label{fig:open_rewardstep}
    \end{minipage}%
    \hfill
    \begin{minipage}[t]{0.48\linewidth}
        \centering
        \includegraphics[width=\linewidth]{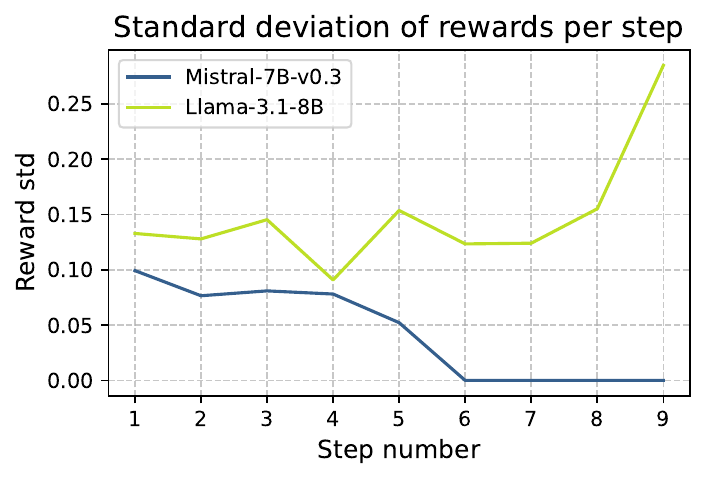}
        \caption{\textbf{Mean standard deviation per step of \decomp with open source models on APPS}.}
        \label{fig:open_stdstep}
    \end{minipage}

\end{figure}

We investigate how different LLMs perform when used with \decomp on 200 competition-level APPS problems, given a sample budget of 30. The groundtruth reward model was used to evaluate correctness, and all models were set to a temperature of 0.8. Due to the challenging nature of the benchmark, open-source models struggled to achieve strong performance independently. However, when paired with \decomp, their performance significantly improved.

\Cref{fig:opensource_token} presents the Pass@token scaling curve for open-source models using \decomp. The results demonstrate that \decomp substantially enhances the capabilities of these models, closing the gap between them and proprietary alternatives.

\Cref{fig:open_actualpart} visualizes the partition frequency of \decomp with different open-source models. Compared to their standalone performance, the use of \decomp led to more structured and effective decomposition, highlighting its adaptability to different architectures.

The mean reward per step is shown in \Cref{fig:open_rewardstep}. Similar to prior findings, we observe that deeper decomposition leads to increasingly higher rewards. Notably, even lower-capacity models benefit from \decomp’s ability to iteratively refine their solutions.

Finally, \Cref{fig:open_stdstep} presents the mean standard deviation per step. With \decomp, the variance in performance is significantly reduced, resulting in more stable and reliable inference.

Overall, these findings emphasize that \decomp is a robust framework capable of enhancing inference performance across diverse LLMs, particularly those with limited standalone capabilities.

\subsection{Ablation on Partition Fraction $\alpha$}
\label{sec:ab_alphafraction}

\begin{figure}[H]
    \centering

    \begin{minipage}[t]{0.48\linewidth}
        \centering
        \includegraphics[width=\linewidth]{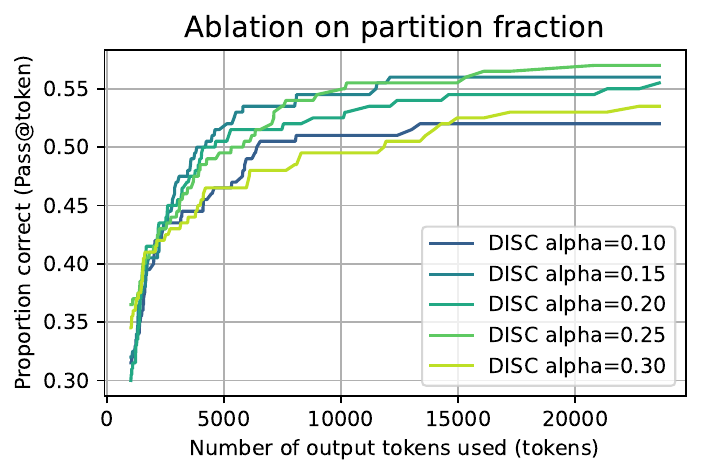}
        \caption{\textbf{Token level comparison of different \decomp splitting fraction $\alpha_0$ on APPS competition level.} $0.15 \le \alpha_0 \le 0.25$ seems to be optimal.}
        \label{fig:alpha_token}
    \end{minipage}%
    \hfill
    \begin{minipage}[t]{0.48\linewidth}
        \centering
        \includegraphics[width=\linewidth]{graphics/alpha_passk.pdf}
        \caption{\footnotesize{\textbf{Effect of partition fraction $\alpha_0$ in \decomp on APPS with gpt-4o-mini.} The range $0.15 \leq \alpha_0 \leq 0.25$ appears optimal.}}
        \label{fig:decomp_ablation_alpha}
    \end{minipage}

\end{figure}

We include additional analysis on the effect of the initial partition fraction $\alpha_0$, which determines the fraction of the best suffix used to propose a new candidate prefix at each iteration. As shown in \Cref{fig:alpha_token,fig:decomp_ablation_alpha}, we observe that performance peaks in the range $0.15 \leq \alpha_0 \leq 0.25$, with both token-level and Pass@k metrics favoring this region.

This behavior aligns with the underlying motivation of \decomp: a smaller $\alpha_0$ results in more conservative candidate proposals—shorter steps that allow for finer-grained refinement. This is beneficial because committing to suboptimal prefixes early in the search can lead to irrevocable errors, especially in greedy search variants. Smaller values of $\alpha_0$ give the algorithm more flexibility to adjust course later on, while still making meaningful progress toward a complete solution. Conversely, large $\alpha_0$ values (e.g., $\alpha_0 > 0.3$) result in overly aggressive expansions that risk committing to noisy or premature completions, leading to reduced accuracy and inefficient use of sampling budget.

Therefore, the optimal range of $\alpha_0$ reflects a balance between exploration and commitment—proposing candidate steps that are informative enough to guide search, yet cautious enough to preserve the ability to refine future decisions. This ablation confirms that adaptive decomposition benefits from conservative, incremental prefix extension when navigating complex reasoning or program synthesis tasks.

\subsection{Reward Distribution}
\label{sec:normal_reward_distribution}

\begin{figure}[H]
    \centering
    \includegraphics[width=0.5\linewidth]{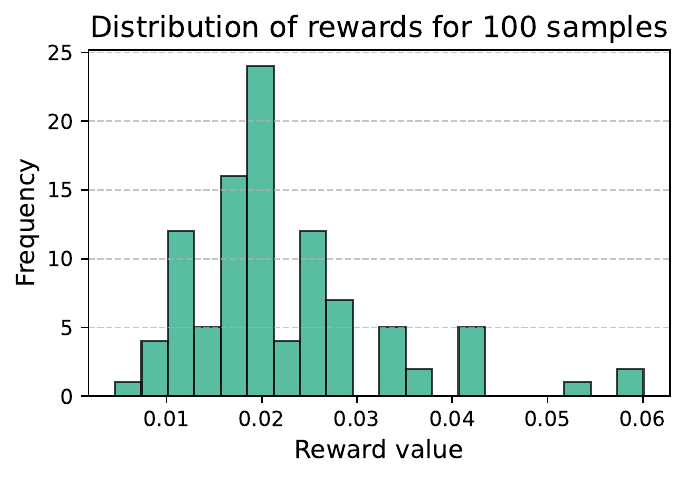}
    \caption{\textbf{Reward distribution for 100 samples on a given MATH500 problem.} The rewards appear to be roughly normal shaped.}
    \label{fig:normal_reward_distribution}
\end{figure}

\noindent
A key assumption in our algorithm is that the distribution of rewards for sampled completions from a given prefix follows a location-scale family, such as the Gaussian distribution. This assumption enables the use of z-scores to estimate the relative quality of candidate prefixes and to guide step acceptance. As shown in \Cref{fig:normal_reward_distribution}, the empirical reward distribution for 100 samples on a representative MATH500 problem appears approximately Gaussian, with a unimodal and symmetric shape. While we do not assume exact normality, this empirical observation supports the use of z-score-based comparisons, as the Gaussian approximation provides a reliable proxy for estimating tail probabilities and potential for reward improvement. We observe similar patterns across other problems and benchmarks, further validating this modeling choice.

%% file: sections/appendix/experiments_extended.tex
\newpage
\section{Main Results Extended}
\subsection{APPS}
\label{sec:apps_extended}
\begin{figure}[ht]
    \centering
    \includegraphics[width=0.5\linewidth]{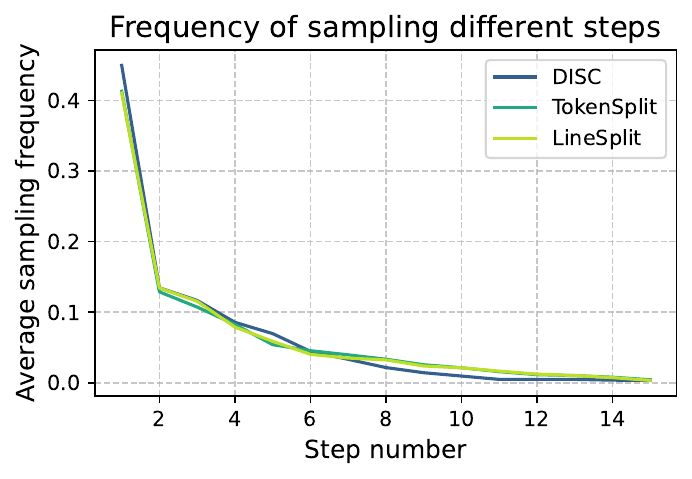}
    \caption{\textbf{Sampling frequency of each step averaged over the problems on APPS with gpt-4o-mini.} \decomp seems to have a slight preferences for spending more compute on earlier found steps.}
    \label{fig:sampling_frequency}
\end{figure}

\begin{figure}[H]
    \centering
    \begin{minipage}{0.49\linewidth}
        \centering
        \includegraphics[width=\linewidth]{graphics/frequency_of_actual_partitons.pdf}
        \label{fig:actual_part_freq}
    \end{minipage}
    \hfill
    \begin{minipage}{0.49\linewidth}
        \centering
        \includegraphics[width=\linewidth]{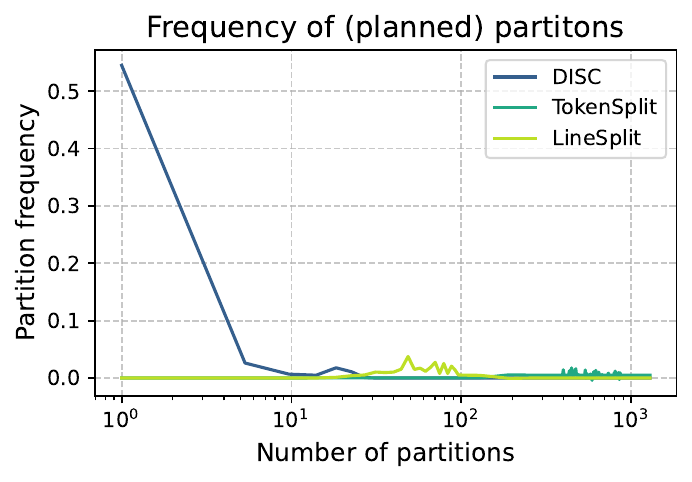}
        \label{fig:planned_part_freq}
    \end{minipage}
 \vspace{-0.7cm}
    \caption{\textbf{Comparison of actual and planned partitions on APPS.}  
    \decomp outperforms other methods with fewer partitions by efficiently identifying critical steps. Unlike token and line split methods, which plan many partitions but search only a subset, \decomp dynamically adjusts partitioning based on budget.}
    \vspace{-0.3cm}
    \label{fig:partition_comparison}
\end{figure}

We provide extended analysis of \decomp’s behavior on the APPS benchmark using \texttt{gpt-4o-mini}. \Cref{fig:sampling_frequency} shows the average sampling frequency per step across all problems. Interestingly, \decomp exhibits a slight preference for allocating more sampling budget to earlier-discovered steps. This behavior reflects the intuition that early steps in a solution often set up the structure for later reasoning, and refining these foundational steps yields more downstream improvements.

\Cref{fig:partition_comparison} compares the number of \textit{actual} partitions (i.e., accepted and committed steps) with the number of \textit{planned} partitions (i.e., proposed candidate splits) across decomposition methods. Token-level and line-based methods typically predefine many steps but only explore a subset due to budget constraints, leading to inefficient allocation. In contrast, \decomp adaptively proposes and accepts partitions in real time based on sample statistics and reward feedback, resulting in significantly fewer actual partitions while maintaining or exceeding performance. This efficiency demonstrates \decomp’s ability to identify and focus on high-impact steps, avoiding wasted compute on unnecessary refinements and enabling stronger performance within the same sampling budget.

\clearpage
\subsection{Additional Examples}
Below is another computed decomposition for the same problem as the one shown in the main text:

\begin{tcolorbox}[colback=white, title=\decomp example decomposition (Lagrange method), colframe=gray]
\textcolor{medium}{\string[ Let \( x \) be the length and \( y \) be the width of the rectangle. Since the perimeter is 24 inches, we have \( 2x + 2y = 24 \Rightarrow x + y = 12\). \string]}
\textcolor{high}{\string[ Therefore, \string]}
\textcolor{low}{\string[ we must maximize the area \( xy \) subject to this constraint. \string]}

\textcolor{low}{\string[ We use the method of Lagrange multipliers. The Lagrangian is}
\[
\textcolor{low}{\mathcal{L}(x, y, \lambda) = xy - \lambda(x + y - 12).}
\]
\textcolor{low}{We differentiate with respect to \( x, y, \) and \( \lambda \) to obtain}
\[
\textcolor{low}{\frac{\partial \mathcal{L}}{\partial x} = y - \lambda = 0, \qquad \frac{\partial \mathcal{L}}{\partial y} = x - \lambda = 0, \qquad \frac{\partial \mathcal{L}}{\partial \lambda} = x + y - 12 = 0.}
\]
\textcolor{low}{We find that \( x = y = \lambda \), so \( x = y = \frac{12}{2} = 6. \) Therefore, the maximum area of the rectangle is \( 6 \cdot 6 = \boxed{36}\). \string]}
\end{tcolorbox}



\decomp again highlights how initial constraint formulation is of medium importance—providing essential structure but not representing a reasoning fork. The token \textbf{"Therefore,"} is marked as high-importance, capturing a critical conceptual transition from constraint setup to the optimization approach. Notably, the bulk of the mathematical machinery involving partial derivatives and substitutions receives \textbf{low importance}, consistent with the idea that such computations are largely mechanical once the decision to use Lagrange multipliers is made. Interestingly, the final step involving the boxed answer is also marked low, suggesting \decomp allocates minimal compute here once earlier reasoning is settled—underscoring the idea that key inference pivots lie upstream in the decision flow.



\subsection{APPS with Self-generated Validation Tests}
\label{sec:val_tests}
We examine \decomp performance on APPS when using self-generated validation tests. All methods utilized the same set of self-generated validation tests to ensure fair comparisons. Each problem received 5-10 validation tests, with the exact number determined dynamically by the LLM. We evaluated a subset of 100 APPS problems, generating samples until the sample budget was exhausted or a correct solution was found.

\begin{figure}[H]
    \centering

    \begin{minipage}[t]{0.48\linewidth}
        \centering
        \includegraphics[width=\linewidth]{graphics/val_passk_scaling.pdf}
        \caption{\textbf{Pass@k on APPS with gpt-4o-mini using self-generated validation tests.}}
        \label{fig:passk_val_selfgen}
    \end{minipage}%
    \hfill
    \begin{minipage}[t]{0.48\linewidth}
        \centering
        \includegraphics[width=\linewidth]{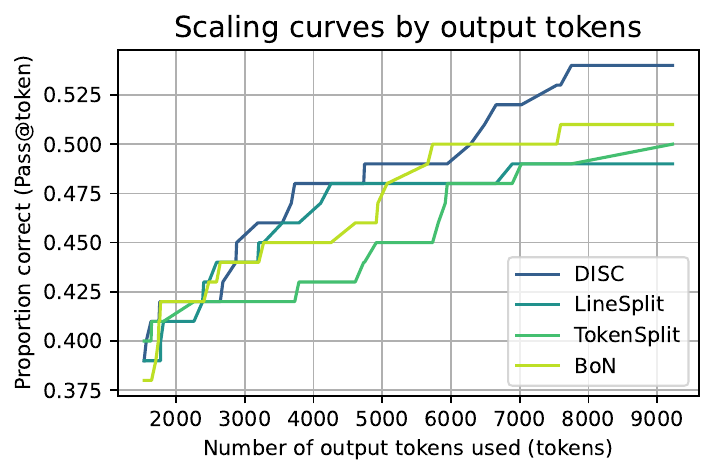}
        \caption{\textbf{Token level comparison of different decomposition methods on APPS with gpt-4o-mini and self-generated validation tests.} \decomp still scales better than other methods in this setting, albeit at a lower rate.}
        \label{fig:val_token}
    \end{minipage}

    \vspace{0.5cm}

    \begin{minipage}[t]{0.48\linewidth}
        \centering
        \includegraphics[width=\linewidth]{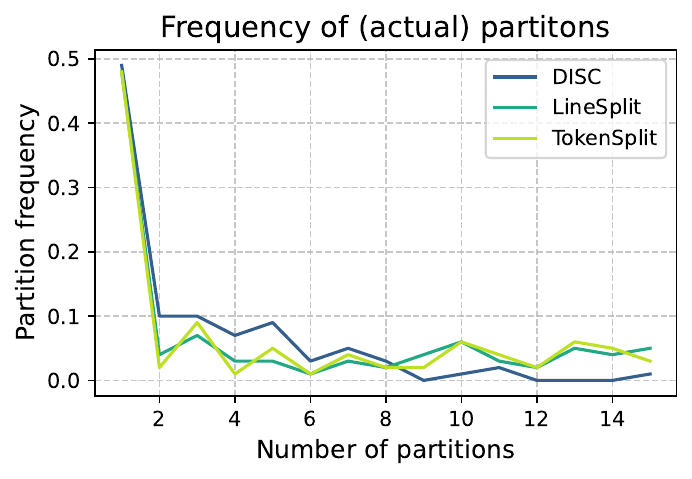}
        \caption{\textbf{Actual partition frequency of different decomposition methods on APPS with gpt-4o-mini and self-generated validation tests.}}
        \label{fig:val_actualpart}
    \end{minipage}%
    \hfill
    \begin{minipage}[t]{0.48\linewidth}
        \centering
        \includegraphics[width=\linewidth]{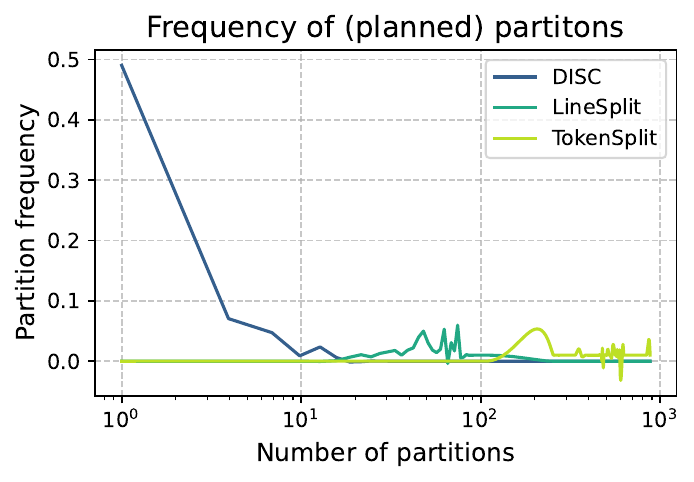}
        \caption{\textbf{Planned partition frequency of different decomposition methods on APPS with gpt-4o-mini and self-generated validation tests.}}
        \label{fig:val_plannedpart}
    \end{minipage}

    \vspace{0.5cm}

    \begin{minipage}[t]{0.48\linewidth}
        \centering
        \includegraphics[width=\linewidth]{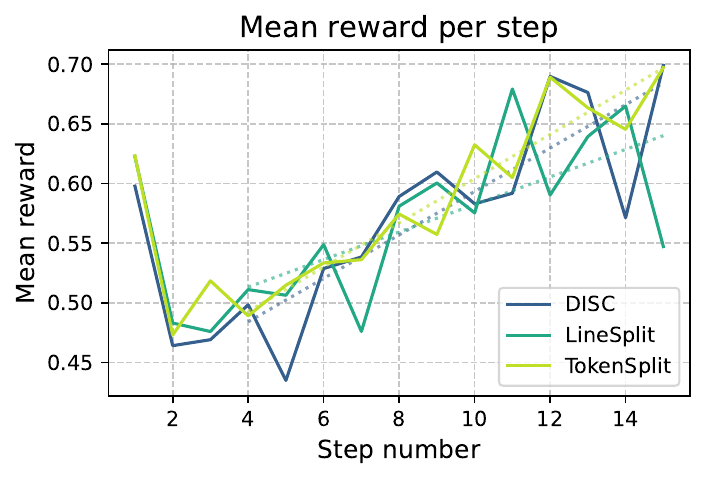}
        \caption{\textbf{Mean reward per step of different decomposition methods on APPS with gpt-4o-mini and self-generated validation tests.}}
        \label{fig:val_rewardstep}
    \end{minipage}%
    \hfill
    \begin{minipage}[t]{0.48\linewidth}
        \centering
        \includegraphics[width=\linewidth]{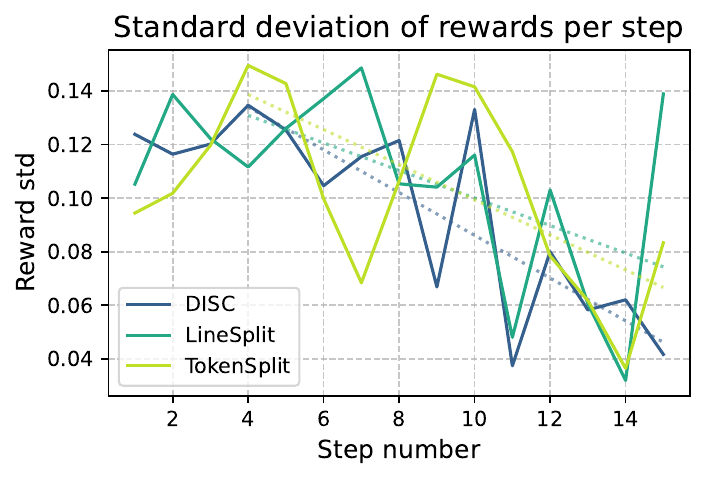}
        \caption{\textbf{Mean standard deviation different decomposition methods on APPS with gpt-4o-mini and self-generated validation tests.}}
        \label{fig:val_stdstep}
    \end{minipage}

\end{figure}

\Cref{fig:val_token} illustrates the Pass@token scaling curve, showing that \decomp maintains strong scaling performance in this setting, though at a slightly lower rate compared to ground-truth verification.

\Cref{fig:val_actualpart} and \Cref{fig:val_plannedpart} compare actual and planned partition frequencies, respectively. The results indicate that \decomp continues to make structured decompositions even with self-generated validation, preserving its efficiency.

The mean reward per step, shown in \Cref{fig:val_rewardstep}, follows a similar trend as in previous experiments, reinforcing that \decomp effectively allocates compute resources for iterative refinement.

Lastly, \Cref{fig:val_stdstep} demonstrates that \decomp maintains lower standard deviations in performance, indicating stable quality improvements across steps.

\subsection{MATH500}
\label{sec:math500_extended}

Completions for MATH500 include both the reasoning steps and the final answer. Since MATH500 contains more problems than APPS200 and MATH problems tend to be relatively easier, solution quality saturates quickly. Therefore, we use a lower sample budget of 10 for these experiments. 

\begin{figure}[H]
    \centering

    \begin{minipage}[t]{0.48\linewidth}
        \centering
        \includegraphics[width=\linewidth]{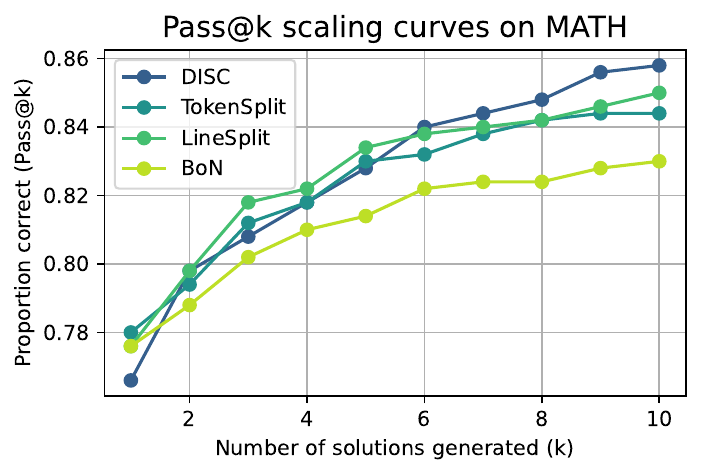}
        \caption{\textbf{Pass@k performance comparison for different decomposition methods on MATH500.} \decomp consistently outperforms BoN across different sampling budgets.}
        \label{fig:math_passk}
    \end{minipage}%
    \hfill
    \begin{minipage}[t]{0.48\linewidth}
        \centering
        \includegraphics[width=\linewidth]{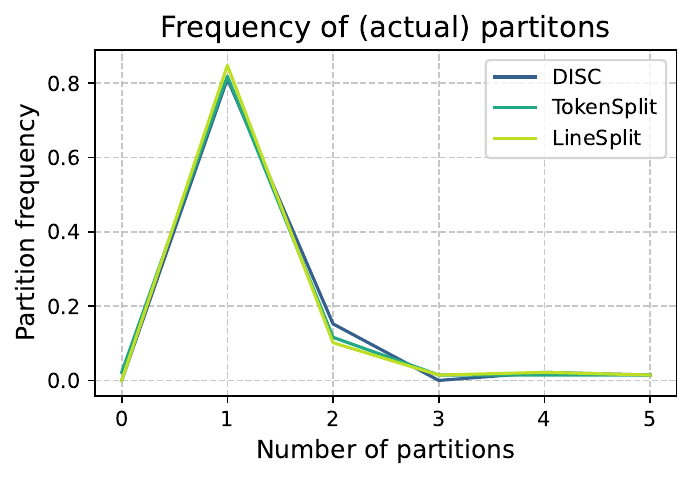}
        \caption{\textbf{Observed partition frequency of different decomposition methods on MATH500.} \decomp effectively segments problems into meaningful subcomponents.}
        \label{fig:math_actualpart}
    \end{minipage}

    \vspace{0.5cm}

    \begin{minipage}[t]{0.48\linewidth}
        \centering
        \includegraphics[width=\linewidth]{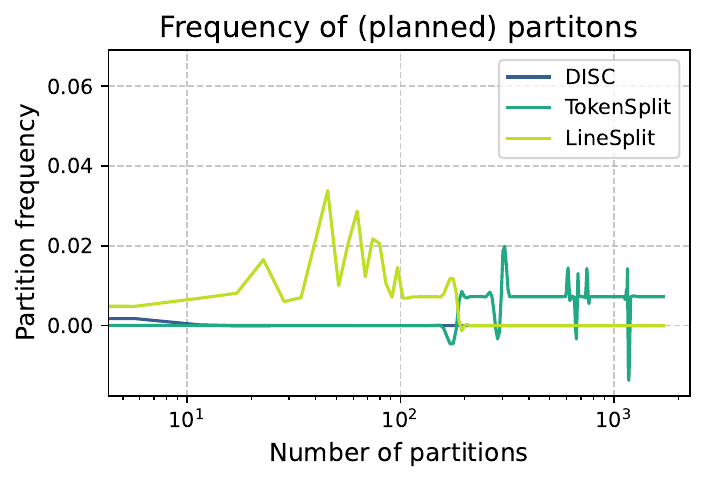}
        \caption{\textbf{Planned partitioning strategy of different decomposition methods on MATH500.} \decomp’s structured approach leads to more efficient problem breakdowns.}
        \label{fig:math_plannedpart}
    \end{minipage}%
    \hfill
    \begin{minipage}[t]{0.48\linewidth}
        \centering
        \includegraphics[width=\linewidth]{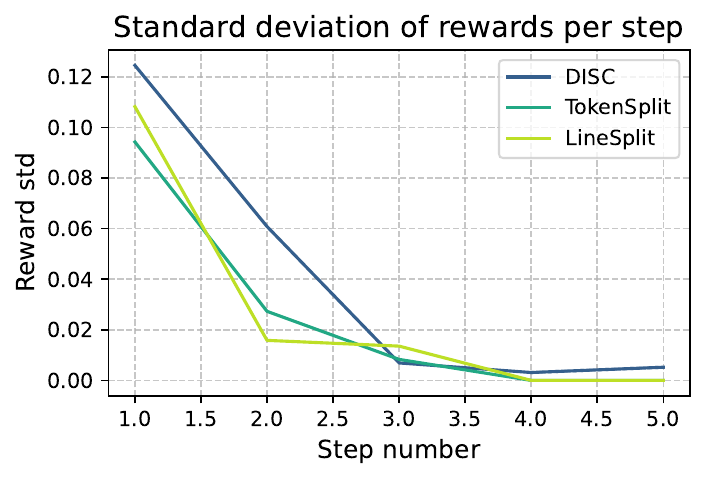}
        \caption{\textbf{Mean standard deviation per step for different decomposition methods on MATH500.} Lower variance in \decomp suggests more stable and reliable problem-solving steps.}
        \label{fig:math_stdstep}
    \end{minipage}

\end{figure}

\Cref{fig:math_passk} presents the Pass@k performance for different decomposition methods on MATH500. We observe that all decomposition-based approaches achieve similar Pass@k performance, consistently outperforming BoN. This indicates that the structured nature of MATH problems allows multiple decomposition strategies to be effective.

Despite similar Pass@k results, the true advantage of \decomp lies in its token efficiency, as shown in \Cref{fig:token_comparison}. \decomp significantly reduces the number of tokens required to reach correct solutions compared to alternative methods, demonstrating its ability to allocate computational effort efficiently in mathematical reasoning tasks.

Additionally, we analyze the partitioning behavior of \decomp on MATH500. \Cref{fig:math_actualpart} illustrates the actual partition frequency for different decomposition methods. The planned partitioning behavior, shown in \Cref{fig:math_plannedpart}, further highlights how \decomp effectively balances exploration and refinement.

Finally, we present the mean standard deviation per step in \Cref{fig:math_stdstep}. Lower variance suggests that \decomp produces more stable and reliable decompositions over multiple runs, reinforcing its robustness in both mathematical and program synthesis domains.

\subsection{LiveCodeBench}
\label{sec:livecodebench_extended}

We evaluate \decomp on LiveCodeBench, a benchmark designed for code generation tasks with a focus on real-world software development challenges. LiveCodeBench presents a unique set of problems requiring both reasoning and structured decomposition, making it a suitable testbed for evaluating \decomp's ability to refine and improve intermediate steps.

\begin{figure}[H]
    \centering

    \begin{minipage}[t]{0.48\linewidth}
        \centering
        \includegraphics[width=\linewidth]{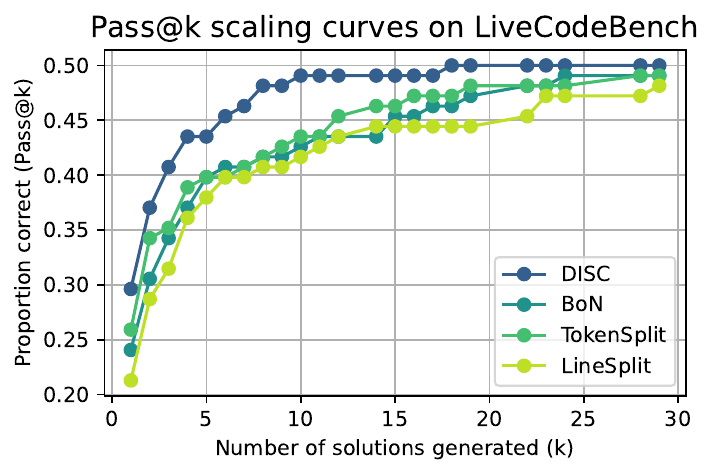}
        \caption{\textbf{Pass@k performance comparison for different decomposition methods on LiveCodeBench.} \decomp consistently outperforms other methods in structured problem refinement.}
        \label{fig:live_passk}
    \end{minipage}%
    \hfill
    \begin{minipage}[t]{0.48\linewidth}
        \centering
        \includegraphics[width=\linewidth]{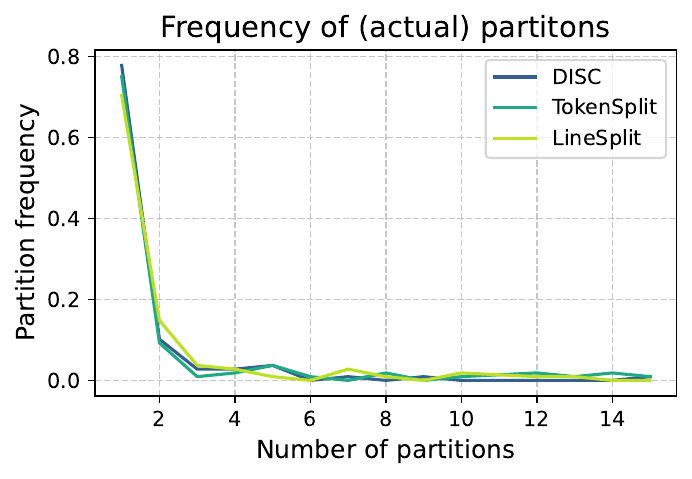}
        \caption{\textbf{Observed partition frequency of different decomposition methods on LiveCodeBench.} \decomp effectively balances problem segmentation while avoiding excessive partitioning.}
        \label{fig:live_actualpart}
    \end{minipage}

    \vspace{0.5cm}

    \begin{minipage}[t]{0.48\linewidth}
        \centering
        \includegraphics[width=\linewidth]{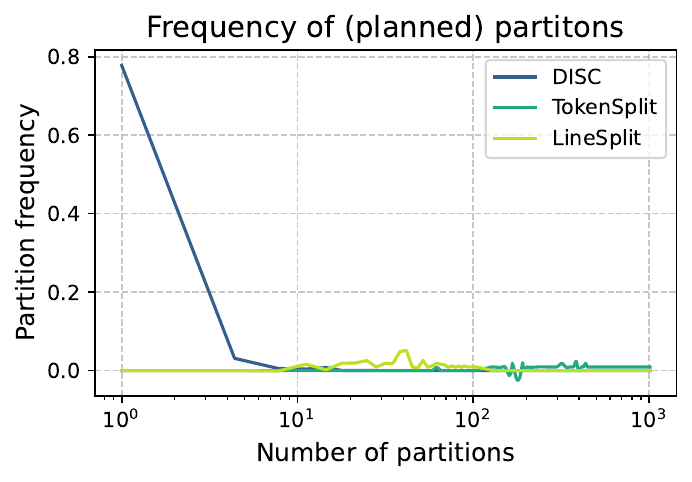}
        \caption{\textbf{Planned partitioning strategy of different decomposition methods on LiveCodeBench.} \decomp dynamically adapts its partitioning to optimize search efficiency.}
        \label{fig:live_plannedpart}
    \end{minipage}%
    \hfill
    \begin{minipage}[t]{0.48\linewidth}
        \centering
        \includegraphics[width=\linewidth]{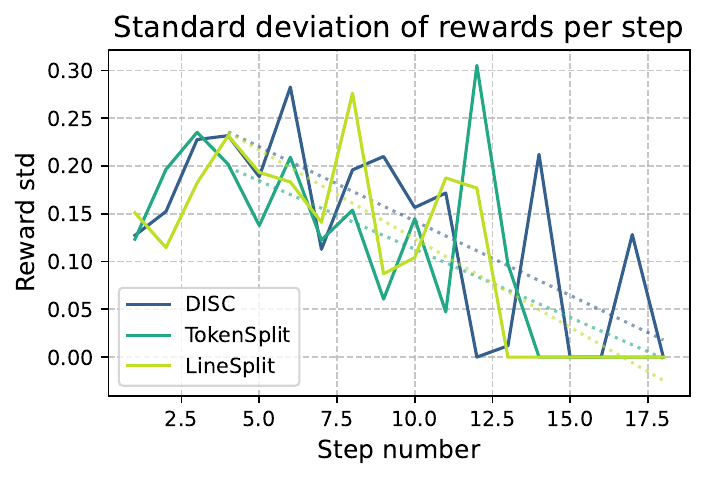}
        \caption{\textbf{Mean standard deviation per step for different decomposition methods on LiveCodeBench.} Lower variance in \decomp suggests more stable and reliable problem-solving steps.}
        \label{fig:live_stdstep}
    \end{minipage}

\end{figure}

\Cref{fig:live_passk} shows the Pass@k comparison of different decomposition methods on LiveCodeBench. \decomp consistently scales better than other decomposition methods, highlighting its ability to refine intermediate steps more effectively in complex coding scenarios.

\Cref{fig:live_actualpart} illustrates the observed partition frequency of different decomposition methods. The structured approach of \decomp results in well-balanced decomposition across steps, reducing unnecessary partitioning while maintaining sufficient granularity for improved solution refinement.

\Cref{fig:live_plannedpart} displays the planned partition frequency across methods. \decomp dynamically determines the most effective partitions based on the evolving problem state, leading to more targeted and efficient decompositions.

Finally, \Cref{fig:live_stdstep} presents the mean standard deviation per step across decomposition methods. Lower variance in \decomp suggests that it produces more stable and reliable decompositions, reinforcing its robustness for solving LiveCodeBench problems.

\subsection{Computational Overhead}
\label{sec:computational_overhead}

We analyze the computational overhead of \decomp by measuring the runtime breakdown between actual LLM token generation and all auxiliary operations (e.g., z-score computation, candidate prefix management, and recursive decomposition logic). Since token generation is the dominant cost in LLM inference, reporting \textit{relative percentages} of runtime—rather than absolute wall-clock time—offers a more meaningful and consistent comparison. Absolute runtime can vary significantly across hardware configurations, backend optimizations, or batching strategies, whereas percentage-based measurements isolate algorithmic overhead from infrastructure-dependent variance.

As shown in \Cref{fig:overhead}, the proportion of runtime \decomp spends on LLM token generation versus auxiliary logic closely matches that of baseline methods such as BoN and LineSplit. This indicates that \decomp introduces negligible additional overhead despite its dynamic behavior and internal scoring computations. In practice, more than 90\% of total runtime is still dominated by LLM token sampling across all methods.

Furthermore, because \decomp achieves better performance with fewer tokens (i.e., improved token efficiency), its effective runtime per successful solution is actually \textit{lower} than that of less efficient baselines. This combination—minimal added overhead and superior token scaling—makes \decomp a highly practical method for real-world inference settings, where total compute cost is often tightly constrained.

%% file: sections/appendix/limitations.tex
\newpage
\section{Limitations}
\label{app:limitations}

While DISC demonstrates strong empirical performance and generality across tasks and models, several limitations merit discussion.

\paragraph{Reliance on LLM-Generated Test Validation.}
Our evaluation framework employs \emph{self-generated test validation}, where the LLM is prompted to produce its own unit tests for a given code generation task, and these tests are used as a proxy reward model. This approach enables scalable evaluation in the absence of ground-truth test cases, but its effectiveness depends critically on the quality of the generated tests. In particular:
\begin{itemize}
    \item The LLM’s ability to produce meaningful and comprehensive tests directly impacts evaluation fidelity. Poor coverage or semantically shallow tests may overestimate correctness.
    \item The framework assumes that most generated tests are correct and executable. For complex or underspecified tasks, this assumption may fail, leading to false positives or noisy reward signals.
    \item Since generated tests may not align with reference solutions, cross-method comparison can become inconsistent.
\end{itemize}
These issues are consistent with observations in prior work such as \emph{CodeT}~\cite{chen2022codet}, which highlights tradeoffs between scalability and reliability in test-based self-evaluation.

\paragraph{Dependence on Reward Model Availability.}
DISC requires access to a scalar reward signal to guide step-wise decomposition. While tasks like code generation or math reasoning naturally provide verifiers (e.g., test cases or numerical checks), applying DISC to domains lacking explicit reward signals necessitates constructing auxiliary reward models or LLM-based critics, which introduces additional complexity and potential bias.

\paragraph{Scope Limited to Single-Turn Generation.}
The current formulation of DISC is designed for single-pass generation tasks. It does not yet handle multi-turn or interactive settings where intermediate reasoning steps can elicit feedback or modify the problem context. Extending DISC to such interactive regimes remains an important direction for future work.

\paragraph{Reduced Benefit in Trivial or Non-Compositional Tasks.}
DISC allocates computational budget adaptively across reasoning steps. When a task is trivially solvable or lacks compositional structure—such that early reasoning does not meaningfully constrain the final outcome—the gains from decomposition and step-wise control diminish.

Despite these limitations, we believe that DISC provides a strong foundation for adaptive, reward-driven reasoning. Future extensions could integrate learned verifiers, support multi-turn interaction, and explore joint reasoning--evaluation co-evolution.

%% file: sections/appendix/search.tex
\newpage
\section{Search and Scaling}
\label{sec:search_extended}

\subsection{\decomp Plug-and-Play Search}

\begin{algorithm}[h]
\caption{\decomp: Decomposition for Plug-and-Play Search}
\label{alg:disc_expansion}
\begin{algorithmic}[1]
\Require LLM $\pi$, Reward model $R$, prompt $x$, initial partition fraction $\alpha_0$, negative binomial threshold $\sigma$, total budget $N$, current budget $n$

\Statex

\State \textbf{function} \textsc{ExpandNode}(parent prefix $\boldsymbol{p}_b$, parent z score $z_b$, suffix to child $\boldsymbol{s}_c$)
\State \quad $\alpha = \alpha_0$
\State \quad \textbf{while } $n < N$ \textbf{ do}
\State \quad \quad $\boldsymbol{p}_c = \boldsymbol{p}_b \cdot \textup{split}(\boldsymbol{s}_c, \alpha)$
\State \quad \quad $Y_c = $\textsc{MakeChildren}($\boldsymbol{p}_c$) 
\State \quad \quad $z_c = (\text{max}(Y_c) - \text{mean}(Y_c))/ \text{std}(Y_c)$
\State \quad \quad $n \leftarrow n+M$
\State \quad \quad \textbf{if } $z_c < z_b$ \textbf{ then}
\State \quad \quad \quad \textbf{break} 
\State \quad \quad \textbf{else }
\State \quad \quad \quad $\alpha \leftarrow \alpha_0 \alpha$ \; 
\State \quad \quad \textbf{end if}
\State \quad \textbf{end while}
\State \quad \textbf{return} $\boldsymbol{p}_c$, $Y_c$
\Statex

\State \textbf{function} \textsc{MakeChildren}(parent prefix $\boldsymbol{p}_b$)
\State \quad $Y = \{ (\boldsymbol{p}_b \cdot \boldsymbol{s}^i, R(\boldsymbol{s}^i)) \mid \boldsymbol{s}^i \sim \pi(\cdot | \boldsymbol{p}_b) \}_{i=1}^{M}$ 
\State \quad \textbf{return} Y

\end{algorithmic}
\end{algorithm}

\subsection{Monte Carlo Tree Search (MCTS)}
\label{sec:mcts}

Monte Carlo Tree Search (MCTS) is a widely used algorithm for sequential decision-making in large search spaces, particularly in applications such as \emph{game playing, planning, and inference scaling}. The algorithm builds a search tree incrementally by simulating different sequences of actions and updating estimates of state quality. A key advantage of MCTS is its ability to balance \emph{exploration} (discovering new states) and \emph{exploitation} (refining promising ones) using a data-driven search process. The MCTS pipeline consists of four fundamental steps: \emph{selection, expansion, simulation, and backpropagation}.

\subsubsection{Selection}
Starting from the root node representing the current state $\boldsymbol{s}$, MCTS iteratively traverses the search tree by selecting child nodes based on a \emph{selection policy}. The most commonly used selection criterion is the \emph{Upper Confidence Bound for Trees (UCT)}, which balances exploration and exploitation:
\begin{equation}
    \label{eq:UCT_mcts}
    UCT(\boldsymbol{s}, \boldsymbol{d}) = \hat{Q}(\boldsymbol{s}, \boldsymbol{d}) + c \sqrt{\frac{\ln \left(\sum_{\boldsymbol{b}} n(\boldsymbol{s}, \boldsymbol{b})\right)}{n(\boldsymbol{s}, \boldsymbol{d})}},
\end{equation}
where $\hat{Q}(\boldsymbol{s}, \boldsymbol{d})$ represents the estimated value of selecting action $\boldsymbol{d}$ from state $\boldsymbol{s}$, $n(\boldsymbol{s}, \boldsymbol{d})$ is the visit count for this action, and $c$ is a hyperparameter controlling the trade-off between exploring new actions and favoring those with high past rewards.

\subsubsection{Expansion}
Once a leaf node (a previously unexplored state) is reached, the algorithm expands the tree by \emph{adding one or more new nodes}. These new nodes represent potential future states $\boldsymbol{s}'$ generated by sampling an action $\boldsymbol{d}$ from a predefined policy. This step broadens the search space and allows MCTS to evaluate new possibilities.

\subsubsection{Simulation}
Following expansion, the algorithm conducts a \emph{simulation} (or rollout) from the newly added state. This step involves generating a sequence of actions according to a predefined policy until reaching a terminal state or an evaluation horizon. The outcome of the simulation, denoted as $v(\boldsymbol{s}')$, provides an estimate of the quality of the new state. Depending on the application, this could represent a \emph{game result, an optimization score, or an inference accuracy metric}.

\subsubsection{Backpropagation}
The final step involves \emph{propagating the results of the simulation back up the search tree} to refine the estimated values of prior states and actions. Each node along the trajectory $\tau = [\boldsymbol{s}_0, \boldsymbol{d}_1, \boldsymbol{s}_2, \dots, \boldsymbol{s}_{-1}]$ is updated iteratively:
\begin{equation}
    \label{eq:backprop_mcts}
    \hat{Q}(\boldsymbol{s}_i, \boldsymbol{d}_{i+1})^{(t+1)} \leftarrow (1-\alpha_n) \hat{Q}(\boldsymbol{s}_i, \boldsymbol{d}_{i+1})^{(t)} + \alpha_n \max\{\hat{Q}(\boldsymbol{s}_i, \boldsymbol{d}_{i+1})^{(t)}, \hat{Q}(\boldsymbol{s}_{i+1}, \boldsymbol{d}_{i+2})^{(t+1)}\},
\end{equation}
where $\alpha_n$ is a learning rate that depends on the visit count, and the maximum function ensures that the best-performing trajectories are emphasized.

MCTS has been widely adopted in inference scaling techniques due to its ability to \emph{efficiently allocate computational resources}, focusing more on \emph{high-reward states} while avoiding unnecessary exploration of unpromising regions. In later sections, we explore how MCTS can be combined with \emph{dynamic decomposition} to further optimize inference scaling.

\subsubsection{Combining Dynamic Decomposition with MCTS}
MCTS can be enhanced by integrating \emph{dynamic decomposition}, where each node in the search tree represents a decomposition of the problem into steps. Instead of treating states as atomic decisions, we recursively decompose reasoning steps, dynamically adjusting granularity based on difficulty.

In this framework:
\begin{itemize}
    \item Each node in the MCTS tree represents a partial decomposition of the problem, with child nodes corresponding to alternative step partitions.
    \item Branching occurs by generating candidate next steps using dynamic decomposition, allowing finer steps for complex regions while maintaining efficiency for simpler ones.
    \item The selection step prioritizes nodes that represent more promising decompositions, dynamically refining challenging areas through recursive subdivision.
    \item The backpropagation step ensures that decompositions leading to high-quality solutions are reinforced, helping the search tree converge toward optimal inference paths.
\end{itemize}
By integrating dynamic decomposition with MCTS, we efficiently allocate compute to the most critical reasoning steps, improving inference quality while maintaining computational efficiency.

\subsection{Beam Search}
\label{sec:beam_search}

Beam search is a heuristic search algorithm commonly used in inference tasks where computational efficiency is a priority. Unlike exhaustive search methods, beam search maintains only the top $k$ best candidates at each step, making it an effective strategy for structured prediction problems and sequential decision-making.

At each iteration:
\begin{itemize}
    \item The algorithm selects the $k$ most promising partitions from the previous step based on an evaluation metric.
    \item Each selected partition is expanded by generating possible next-step samples.
    \item The newly generated partitions are ranked, and only the top $k$ candidates are retained for the next iteration.
    \item This process continues until a stopping criterion is met, such as reaching a predefined depth or finding a sufficiently high-quality solution.
\end{itemize}

Beam search provides a computationally efficient way to explore structured solution spaces while maintaining high-quality search trajectories. By integrating beam search with dynamic decomposition, we ensure that inference computation is allocated efficiently, focusing on the most promising reasoning paths at each step.

\subsection{Additional Results and Analysis}

\begin{figure}[H]
    \centering

    \begin{minipage}[t]{0.48\linewidth}
        \centering
        \includegraphics[width=\linewidth]{graphics/search_passk.pdf}
        \caption{\textbf{Pass@k on APPS with gpt-4o-mini using different search methods.} MCTS scales best, followed by greedy search, followed by beam search.}
        \label{fig:passk_search_methods}
    \end{minipage}%
    \hfill
    \begin{minipage}[t]{0.48\linewidth}
        \centering
        \includegraphics[width=\linewidth]{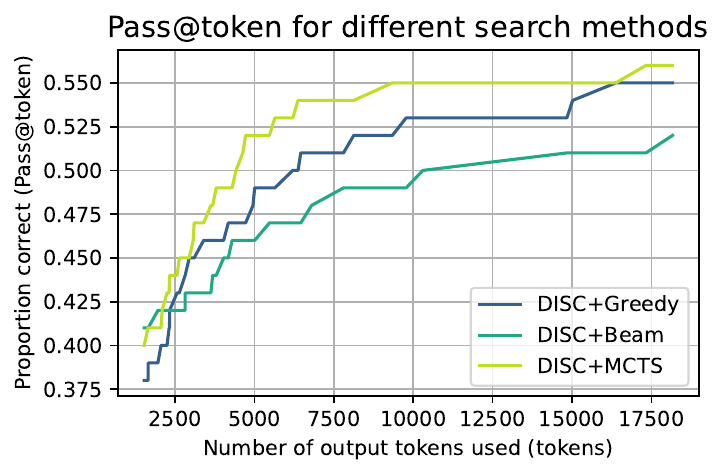}
        \caption{\textbf{Token level comparison of different decomposition search methods combined with \decomp on APPS with gpt-4o-mini.} MCTS scales best, followed by greedy search, followed by beam search.}
        \label{fig:search_token}
    \end{minipage}

    \vspace{0.5cm}

    \begin{minipage}[t]{0.48\linewidth}
        \centering
        \includegraphics[width=\linewidth]{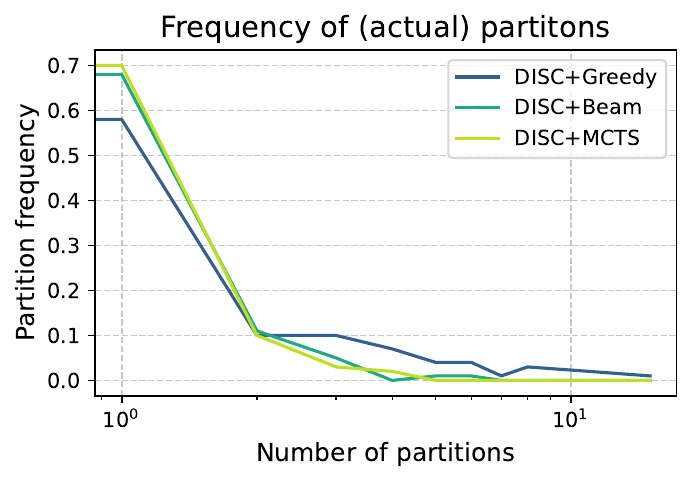}
        \caption{\textbf{Actual partition frequency of different decomposition search methods combined with \decomp on APPS with gpt-4o-mini.} Greedy is able to search to higher depths given the same sampling budget.}
        \label{fig:search_actualpart}
    \end{minipage}%
    \hfill
    \begin{minipage}[t]{0.48\linewidth}
        \centering
        \includegraphics[width=\linewidth]{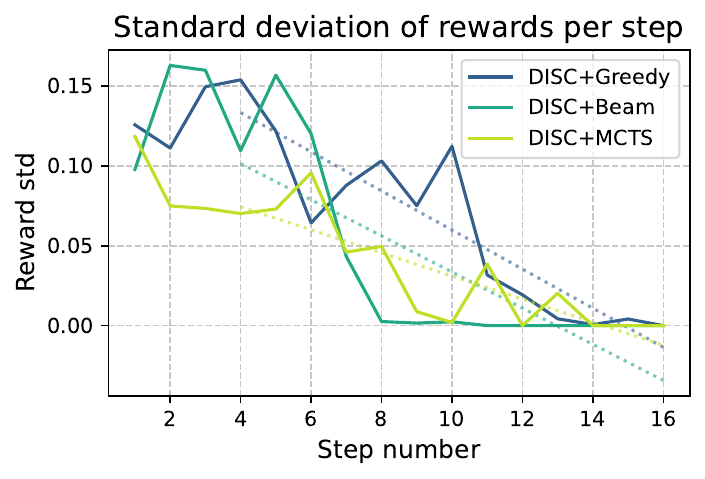}
        \caption{\textbf{Mean standard deviation of different decomposition search methods combined with \decomp on APPS with gpt-4o-mini.} All search methods display decreasing standard deviation with search depth.}
        \label{fig:search_stdstep}
    \end{minipage}

\end{figure}

Experiments comparing different search methods were conducted on a 100-problem subset of the APPS dataset (first 100 problems) using GPT-4o-mini. All methods used a temperature of 0.2, with $\alpha=0.15$, Q priority metric, and $\sigma=1.0$.

\textbf{Token-level comparison:} As shown in Figure \ref{fig:search_token}, MCTS scales best among the tested methods, demonstrating superior efficiency in identifying promising partitions. Greedy search follows closely, while beam search exhibits the slowest scaling.

\textbf{Partition frequency analysis:} Figure \ref{fig:search_actualpart} reveals that greedy search explores to greater depths within the same sampling budget. This suggests that greedy search prioritizes deep refinements, whereas MCTS and beam search balance depth with breadth.

\textbf{Step variance analysis:} Figure \ref{fig:search_stdstep} illustrates that all search methods display decreasing standard deviation with increasing search depth. This trend indicates that deeper searches converge towards stable, high-quality partitions, reinforcing the benefits of dynamic decomposition.

These results highlight the trade-offs between search methods: MCTS offers robust exploration-exploitation balance, greedy search favors depth-first refinement, and beam search provides a structured yet computationally constrained approach. The integration of dynamic decomposition further enhances these search strategies by adaptively allocating computational resources to critical reasoning steps.



%% file: sections/appendix/analysis.tex
\newpage
\section{Theoretical analysis}
\label{sec:theory}

\subsection{Main Theorem}
We begin by introducing notation to characterize the distribution of rewards obtained when sampling completions from a given prefix $\boldsymbol{p}$. 

Let $R_{\boldsymbol{p}}$ denote the random variable of the reward of a completion sampled from the LLM policy $\pi$ conditioned on prefix $\boldsymbol{p}$:
\[
R_{\boldsymbol{p}} := R(\boldsymbol{p} \cdot \boldsymbol{s}), \quad \text{where } \boldsymbol{s} \sim \pi(\cdot \mid \boldsymbol{p}).
\]
Let $F_{\boldsymbol{p}}$, $\mu_{\boldsymbol{p}}$, and $\sigma_{\boldsymbol{p}}$ be the cumulative distribution function (CDF), mean, and standard deviation of $R_{\boldsymbol{p}}$, respectively.

We define a solution $\boldsymbol{y}^*$ as \emph{correct} if it achieves a reward of at least $r^*=1$, i.e., $R(\boldsymbol{y}^*) \ge 1$. Then, the probability that a random sample from $\pi(\cdot \mid \boldsymbol{p})$ yields a correct solution is given by:
\[
\mathbb{P}[R_{\boldsymbol{p}} \ge 1] = 1 - F_{\boldsymbol{p}}(1).
\]

For a given set of $M$ independent samples from $\pi(\cdot \mid \boldsymbol{p})$, let $r^{\max}_{\boldsymbol{p}}$ denote the maximum observed reward among them, i.e., the sample maximum.

We make the following problem assumption: 
\begin{assumption}
    \label{assumption:problem}
    A correct solution exists in the support of the policy $\pi(\boldsymbol{y}^* | \boldsymbol{x}) > 0$ and the length of the correct solution is finite: $|\boldsymbol{y}^*| = K < \infty$. 
\end{assumption}

We make the following assumptions about the policy $\pi$:

\begin{assumption}
    We assume $R_{\boldsymbol{p}}$ belongs to a location-scale family with base CDF $F_R$. 
\end{assumption}

\begin{remark}
    The location-scale family includes common distributions such as the Normal, Cauchy, and Logistic distributions, which are characterized by being affine transformations of a fixed base distribution. Specifically, a random variable $X$ belongs to a location-scale family if it can be written as $X = \mu + \sigma Z$, where $Z \sim F_R$ is a standardized random variable, and $\mu \in \mathbb{R}$ and $\sigma > 0$ are the location and scale parameters, respectively. We observe empirical evidence supporting this modeling choice for $R_{\boldsymbol{p}}$ in \Cref{sec:normal_reward_distribution}.
\end{remark}

\begin{assumption}[Reward distribution converges with prefix]
\label{assumption:reward_variation_convergence}
Consider some base prefix $\boldsymbol{b}$ and some candidate prefix $\boldsymbol{c} = \boldsymbol{b} \cdot \boldsymbol{x}$ for any token sequence $\boldsymbol{x}$. 
Then, $\sigma_{\boldsymbol{c}} \leq \sigma_{\boldsymbol{b}}$. 
Furthermore, if $\boldsymbol{b}$ is a complete solution (it ends in an EOS token) then $\sigma_{\boldsymbol{b}} = 0$. 
\end{assumption}

\begin{remark}
    This assumption reflects the intuition that as more tokens are appended to a prefix, the LLM becomes increasingly committed to a narrower set of likely continuations, leading to lower uncertainty in the reward distribution. In the limit, once a complete solution is formed (i.e., the sequence ends with an EOS token), the reward becomes deterministic, and the variance collapses to zero.

    We emphasize that this assumption is not required for proving the optimality of \decomp. However, the greater the difference between $\sigma_{\boldsymbol{c}} < \sigma_{\boldsymbol{b}}$, the faster the convergence, as \decomp will be able to prune suboptimal paths more confidently with fewer samples.
\end{remark}

\begin{assumption}[Reward distribution converges slowly]
\label{assumption:smooth_convergence}
If \Cref{alg:discwithgreedy} accepts a candidate prefix $\boldsymbol{c}$ over a base prefix $\boldsymbol{b}$, then we assume the standard deviation of the reward distribution under $\boldsymbol{c}$ is a lower bounded: 
\[
\sigma_{\boldsymbol{c}} \geq \left(1 - \frac{\delta \sigma_{\boldsymbol{c}}}{\Delta} \right)\sigma_{\boldsymbol{b}}
\]
where
$\Delta = r^* - r^{\max}_{\boldsymbol{c}}$ is the \textbf{optimality gap} between the true maximum reward $r^*$ and the current best sample under prefix $\boldsymbol{b}$,
and $\delta = \frac{r^{\max}_{\boldsymbol{b}} - \mu_{\boldsymbol{b}}}{\sigma_{\boldsymbol{b}}} - \frac{r^{\max}_{\boldsymbol{c}} - \mu_{\boldsymbol{c}}}{\sigma_{\boldsymbol{c}}}$ is the \textbf{z-score drop}, which quantifies the change in how surprising the best sample is under each distribution.
\end{assumption}

\begin{remark}
This assumption enforces that if we transition from base prefix $\boldsymbol{b}$ to candidate prefix $\boldsymbol{c}$, the standard deviation (spread) of the reward distribution under $\boldsymbol{c}$ cannot shrink too drastically—especially when:
(i) we are still far from the optimal reward (i.e., $\Delta$ is large), and
(ii) the improvement in sample quality is not significant (i.e., $\delta$ is small).
Intuitively, this prevents committing to a sharp but unreliable prefix unless there is strong evidence of progress. 
If $\boldsymbol{c}$ does not yield sufficiently better samples or has low variance, it may hinder further exploration. This assumption ensures that accepted prefixes preserve enough reward diversity to support continued search.
\end{remark}

\begin{remark}
While we list the confidence property~\Cref{assumption:smooth_convergence} as an assumption, it can also be enforced directly as an \textbf{explicit acceptance criterion} in \Cref{alg:discwithgreedy}. Specifically, the algorithm could be modified to accept a candidate prefix $\boldsymbol{c}$ over base $\boldsymbol{b}$ only if both $\delta \ge 0$ and:
$
\left(1 - \frac{\delta \sigma_{\boldsymbol{c}}}{\Delta} \right)\sigma_{\boldsymbol{b}} \le \sigma_{\boldsymbol{c}}.
$
where $\Delta$ can be computed because $r^* = 1$. 
\end{remark}

Combining Assumptions~\ref{assumption:reward_variation_convergence} and~\ref{assumption:smooth_convergence}, we can write: $\frac{\sigma_c}{\sigma_b} \in \left[ \left(1 - \frac{\delta \sigma_{\boldsymbol{c}}}{\Delta} \right), 1 \right]$. This balances exploration with exploitation.

\begin{assumption}[Accurate estimates of $\mu_{\boldsymbol{p}}$ and $\sigma_{\boldsymbol{p}}$ ]
\label{assumption:estimate_reward_dist}
We assume that we sample $R_{\boldsymbol{p}}$ enough times to get accurate estimates of the true mean $\hat{\mu}_{\boldsymbol{p}} \approx \mu_{\boldsymbol{p}}$ and standard deviation $\hat{\sigma}_{\boldsymbol{p}} \approx\sigma_{\boldsymbol{p}}$. 
Because estimates of mean and standard deviation are accurate, the estimates of the z-scores are also accurate.
\end{assumption}




Next we prove that Alg.~\ref{alg:discwithgreedy} will never commit a candidate prefix that with a lower probability of sampling a correct solution than the base. 

\begin{lemma}
\label{lemma:iteration_invariance}
Consider Algorithm~\ref{alg:discwithgreedy}, applied to a base prefix $\boldsymbol{b}$ and a candidate prefix $\boldsymbol{c}$, with corresponding reward distributions $F_{\boldsymbol{b}}$ and $F_{\boldsymbol{c}}$. Then, the probability of sampling a correct suffix from $\pi$ does not decrease after accepting $\boldsymbol{c}$:
\[
\pi(\boldsymbol{s}^* \mid \boldsymbol{b}) \leq \pi(\boldsymbol{s}^* \mid \boldsymbol{c}).
\]
\end{lemma}

\begin{proof}
A candidate prefix $\boldsymbol{c}$ is accepted by Algorithm~\ref{alg:discwithgreedy} if either of the two conditions are true: 
(i) the standardized score (z-score) of $\boldsymbol{c}$ is lower than that of $\boldsymbol{b}$:
$z_c < z_b$ 
or 
(ii) the first $\alpha$ fraction of tokens in the best suffix contains no tokens, and the prefix is accepted by default.
In case (ii): if $\boldsymbol{c}$ is the empty prefix, then $\boldsymbol{c} = \boldsymbol{b}$, and the distributions are unchanged, so the result follows trivially.

It remains to show case (i): if $z_c < z_b$ then, $\pi(\boldsymbol{s}^* \mid \boldsymbol{c}) \geq \pi(\boldsymbol{s}^* \mid \boldsymbol{b})$.

First, we define the z-score change $\delta>0$ as $\delta = \frac{r^{\max}_{\boldsymbol{b}} - \mu_{\boldsymbol{b}}}{\sigma_{\boldsymbol{b}}} - \frac{r^{\max}_{\boldsymbol{c}} - \mu_{\boldsymbol{c}}}{\sigma_{\boldsymbol{c}}}$ and the optimality gap $\Delta > 0$ as $\Delta = r^* - r^{\max}_{\boldsymbol{c}}$. Then, we note that our algorithmic implementation initializes the set of candidate samples with the best solution, i.e. $\text{max}(Y_b) \in Y_c$. Therefore, we have that the sample max of the candidate rewards is greater than or equal to that of the base: $r_{\boldsymbol{c}}^{\max} \geq r_{\boldsymbol{b}}^{\max}$.

Using this information, we can rewrite the z-score inequality: 
\begin{align}
    \frac{r^* - \mu_{\boldsymbol{b}}}{\sigma_{\boldsymbol{b}}} - \frac{r^* - \mu_{\boldsymbol{c}}}{\sigma_{\boldsymbol{c}}} 
    &\ge \left( \delta - \frac{\Delta}{\sigma_{\boldsymbol{c}}} + \frac{\Delta}{\sigma_{\boldsymbol{b}}} \right) 
    \geq 0
\end{align}
where we first plug in the definitions of $\delta$ and $\Delta$, do algebraic manipulation, and then apply assumption~\ref{assumption:smooth_convergence}. The final inequality implies $F_{\boldsymbol{c}}(r^*) < F_{\boldsymbol{b}}(r^*)$, which implies $\pi(\boldsymbol{s}^* \mid \boldsymbol{c}) \geq \pi(\boldsymbol{s}^* \mid \boldsymbol{b})$ because $\mathbb{P}[R_{\boldsymbol{p}} \ge r^*] = 1 - F_{\boldsymbol{p}}(r^*)$ and $F_{\boldsymbol{p}}$ is monotonic in z score. 

\end{proof}

\begin{theorem}[Optimality of \decomp]
\label{thm:optimality_repeat}
Consider Algorithm~\ref{alg:discwithgreedy} applied to a problem input $\boldsymbol{x}$ and assume Assumptions~1, 2, 4, and 5 hold. 
Then, with probability 1, there exists a finite number of accepted prefixes $k > 0$ and algorithm iterations $n>0$ such that the algorithm terminates, and the best solution $\boldsymbol{y}_k$ found is a correct solution: $R(\boldsymbol{y}_k) \geq 1$.
\end{theorem}

\begin{proof}
We define the base prefix $\boldsymbol{b}_k$ as the result of appending $k$ accepted candidate prefixes. We now proceed with the induction.

We first prove by induction that the probability of sampling a correct suffix remains strictly positive after each accepted prefix. 

\textbf{Base Case ($k = 0$)}:  
Initially, $\boldsymbol{b}_0 = \boldsymbol{x}$. By assumption, there exists at least one correct solution $\boldsymbol{y}^* = \boldsymbol{x} \cdot \boldsymbol{s}^*$ such that $\pi(\boldsymbol{s}^* \mid \boldsymbol{x}) > 0$ and $R(\boldsymbol{y}^*) \geq 1$. Hence, there is a non-zero probability of sampling a correct suffix from $\boldsymbol{b}_0$.

\textbf{Inductive Step:}  
Assume that after $k$ accepted prefixes, the base prefix $\boldsymbol{b}_k$ satisfies:
\[
\pi(\boldsymbol{s}^* \mid \boldsymbol{b}_k) > 0 \quad \text{and} \quad R(\boldsymbol{b}_k \cdot \boldsymbol{s}^*) \geq 1
\]
for some correct suffix $\boldsymbol{s}^*$. Let $\boldsymbol{b}_{k+1} = \boldsymbol{b}_k \cdot \boldsymbol{c}$ be the base prefix after appending the $(k+1)$-th accepted candidate prefix $\boldsymbol{c}$. By Lemma~\ref{lemma:iteration_invariance}, the probability of sampling a correct suffix does not decrease:
\[
\pi(\boldsymbol{s}^* \mid \boldsymbol{b}_{k+1}) \geq \pi(\boldsymbol{s}^* \mid \boldsymbol{b}_k) > 0.
\]

Therefore, at every accepted prefix $\boldsymbol{b}_k$, the probability of sampling a correct suffix remains bounded below by some $\varepsilon > 0$, where $\varepsilon \geq \pi(\boldsymbol{y}^* \mid \boldsymbol{x})$.


\textbf{Termination:}  
Let $M_k$ be the number of samples from accepted candidate prefix $\boldsymbol{b}_k$. Since we sample from the accepted candidate prefix $\boldsymbol{b}_k$ at least once ($M_k \ge 1$), samples are independent, and the probability of sampling a correct suffix is at least $\pi(\boldsymbol{y}^* \mid \boldsymbol{x}) > 0$, the probability of not sampling a correct solution after accepting $k$ prefixes is at most
\[
\left(1 - \pi(\boldsymbol{y}^* \mid \boldsymbol{x})\right)^k.
\]
Hence, the probability of never sampling a correct solution after infinitely many accepted prefixes is
\[
\lim_{k \to \infty} \left(1 - \pi(\boldsymbol{y}^* \mid \boldsymbol{x})\right)^k = 0,
\]
which implies that, with probability 1, the algorithm will eventually sample a correct suffix.

Furthermore, this implies that the number of accepted prefixes $k$ before obtaining a correct solution is almost surely finite. This follows from the fact that the number of independent trials before the first success with fixed success probability $p := \pi(\boldsymbol{y}^* \mid \boldsymbol{x}) > 0$ is a geometric random variable, which is finite with probability 1.


Therefore, there exists a finite number of accepted prefixes $k > 0$ such that the algorithm terminates and returns a correct solution $\boldsymbol{y}^*$ satisfying $R(\boldsymbol{y}^*) \geq 1$.

Since the algorithm either accepts a candidate prefix or contracts it by a factor $\alpha \in (0,1)$ until the prefix length is 0, there can only be finitely many contractions before the candidate prefix is rejected or accepted. Therefore the algorithm must accept a candidate prefix in finite number of iterations, and so the total number of algorithm iterations $n$ before finding a correct solution is also finite.

\end{proof}

\begin{remark}
    Unlike Best-of-N (BoN), where the probability of sampling a correct solution remains constant across samples, \decomp dynamically refines the sampling distribution by appending informative prefixes. As established in Lemma~\ref{lemma:iteration_invariance}, the probability of generating a correct solution under \decomp is non-decreasing across iterations. Consequently, \decomp achieves faster convergence to a correct solution in expectation compared to BoN. This theoretical advantage is corroborated by our empirical results in \Cref{fig:token_comparison}, which show that \decomp consistently outperforms BoN in terms of efficiency and scalability with respect to token budget.
\end{remark}

\subsection{A Motivating Example on \decomp}
\label{sec:motivating_example}
We use the Wiener process $W(t)$ as an example where there are intractably many actions and steps. Suppose we start at $t=0$ with $W(0) = 0$. 
At each round $k$, the algorithm can choose one of the two options:
\begin{enumerate}[leftmargin=*,itemsep=0pt,topsep=0pt,parsep=0pt, partopsep=0pt]
    \item samples a trajectory and observe the final value $W(T)$ at time $t=T$, as the reward signal. Denote the whole trajectory as $w_k(\cdot)$.
    \item chooses one trajectory from the previous rounds (denoted as $w_s(t)$ for round $s$), and time $t_0$; then sample a trajectory at $t=t_0$ with $W(t_0)=w_s(t_0)$. Denote the concatenated trajectory as $w_k(\cdot)$ with $w_k(t)=w_s(t)$ when $t\le t_0$.
\end{enumerate}
Note that we are only able to observe the final reward $W(t)$. At any intermediate time $t \in (0, T)$, the current value $W(t)$ is not observable. The goal is to design an algorithm that can reach the highest reward among the $K$ trajectories. Formally speaking, we aim to maximize the maximum:
\begin{align*}
    \max_{k \in K} w_k(T).
\end{align*}
One naive solution is to call option 1 for $K$ times and return the best-of-$K$ reward, each following:
\begin{align*}
    W(T) & \sim \mathcal{N}(0, T).
\end{align*}

Alternatively, suppose there is a promising path $w(\cdot)$ with a high final reward $w(T)=R$.
It is natural to consider starting at some midpoint $\alpha T$ ($0 < \alpha < 1$) and perform more completions to obtain an even higher reward than $R$. 
The reward distribution sampled this way is
\begin{align*}
    W'(T) \sim \mathcal{N}(w(\alpha T) , (1-\alpha)T).
\end{align*}
The remaining question is which $\alpha$ we should choose. One option is to maximize the probability that the newly sampled reward is higher than $R$:
\begin{align*}
    \mathbb{P}(W'(T) > R) = 1 - \Phi\bigg(  
    \frac{R - w(\alpha T)}{\sqrt{(1-\alpha)T}}
    \bigg).
\end{align*}

%% file: neurips_2025.bbl
\begin{thebibliography}{10}

\bibitem{feng2023alphazero}
Xidong Feng, Ziyu Wan, Muning Wen, Stephen~Marcus McAleer, Ying Wen, Weinan Zhang, and Jun Wang.
\newblock Alphazero-like tree-search can guide large language model decoding and training.
\newblock {\em arXiv preprint arXiv:2309.17179}, 2023.

\bibitem{zeng2024scaling}
Zhiyuan Zeng, Qinyuan Cheng, Zhangyue Yin, Bo~Wang, Shimin Li, Yunhua Zhou, Qipeng Guo, Xuanjing Huang, and Xipeng Qiu.
\newblock Scaling of search and learning: A roadmap to reproduce o1 from reinforcement learning perspective, 2024.

\bibitem{wu2024inferencescalinglawsempirical}
Yangzhen Wu, Zhiqing Sun, Shanda Li, Sean Welleck, and Yiming Yang.
\newblock Inference scaling laws: An empirical analysis of compute-optimal inference for problem-solving with language models, 2024.

\bibitem{nori2024medprompto1explorationruntime}
Harsha Nori, Naoto Usuyama, Nicholas King, Scott~Mayer McKinney, Xavier Fernandes, Sheng Zhang, and Eric Horvitz.
\newblock From medprompt to o1: Exploration of run-time strategies for medical challenge problems and beyond, 2024.

\bibitem{snell2024scaling}
Charlie Snell, Jaehoon Lee, Kelvin Xu, and Aviral Kumar.
\newblock Scaling llm test-time compute optimally can be more effective than scaling model parameters.
\newblock {\em arXiv preprint arXiv:2408.03314}, 2024.

\bibitem{brown2024large}
Bradley Brown, Jordan Juravsky, Ryan Ehrlich, Ronald Clark, Quoc~V Le, Christopher R{\'e}, and Azalia Mirhoseini.
\newblock Large language monkeys: Scaling inference compute with repeated sampling.
\newblock {\em arXiv preprint arXiv:2407.21787}, 2024.

\bibitem{gandhi2024stream}
Kanishk Gandhi, Denise Lee, Gabriel Grand, Muxin Liu, Winson Cheng, Archit Sharma, and Noah~D Goodman.
\newblock Stream of search (sos): Learning to search in language.
\newblock {\em arXiv preprint arXiv:2404.03683}, 2024.

\bibitem{lee2025evolvingdeeperllmthinking}
Kuang-Huei Lee, Ian Fischer, Yueh-Hua Wu, Dave Marwood, Shumeet Baluja, Dale Schuurmans, and Xinyun Chen.
\newblock Evolving deeper llm thinking, 2025.

\bibitem{light2024strategist}
Jonathan Light, Min Cai, Weiqin Chen, Guanzhi Wang, Xiusi Chen, Wei Cheng, Yisong Yue, and Ziniu Hu.
\newblock Strategist: Learning strategic skills by llms via bi-level tree search.
\newblock {\em arXiv preprint arXiv:2408.10635}, 2024.

\bibitem{anonymous2025planning}
Evan Wang, Federico Cassano, Catherine Wu, Yunfeng Bai, Will Song, Vaskar Nath, Ziwen Han, Sean Hendryx, Summer Yue, and Hugh Zhang.
\newblock Planning in natural language improves {LLM} search for code generation.
\newblock In {\em The Thirteenth International Conference on Learning Representations}, 2025.

\bibitem{yao2024tree}
Shunyu Yao, Dian Yu, Jeffrey Zhao, Izhak Shafran, Tom Griffiths, Yuan Cao, and Karthik Narasimhan.
\newblock Tree of thoughts: Deliberate problem solving with large language models.
\newblock {\em Advances in Neural Information Processing Systems}, 36, 2024.

\bibitem{zelikman2023parsel}
Eric Zelikman, Qian Huang, Gabriel Poesia, Noah Goodman, and Nick Haber.
\newblock Parsel: Algorithmic reasoning with language models by composing decompositions.
\newblock {\em Advances in Neural Information Processing Systems}, 36:31466--31523, 2023.

\bibitem{zhou2022least}
Denny Zhou, Nathanael Sch{\"a}rli, Le~Hou, Jason Wei, Nathan Scales, Xuezhi Wang, Dale Schuurmans, Claire Cui, Olivier Bousquet, Quoc Le, et~al.
\newblock Least-to-most prompting enables complex reasoning in large language models.
\newblock {\em arXiv preprint arXiv:2205.10625}, 2022.

\bibitem{lin2025criticaltokens}
Zicheng Lin, Tian Liang, Jiahao Xu, Qiuzhi Lin, Xing Wang, Ruilin Luo, Chufan Shi, Siheng Li, Yujiu Yang, and Zhaopeng Tu.
\newblock Critical tokens matter: Token-level contrastive estimation enhances llm's reasoning capability, 2025.

\bibitem{guo2025deepseek}
Daya Guo, Dejian Yang, Haowei Zhang, Junxiao Song, Ruoyu Zhang, Runxin Xu, Qihao Zhu, Shirong Ma, Peiyi Wang, Xiao Bi, et~al.
\newblock Deepseek-r1: Incentivizing reasoning capability in llms via reinforcement learning.
\newblock {\em arXiv preprint arXiv:2501.12948}, 2025.

\bibitem{chen2021evaluating}
Mark Chen, Jerry Tworek, Heewoo Jun, Qiming Yuan, Henrique Ponde De~Oliveira Pinto, Jared Kaplan, Harri Edwards, Yuri Burda, Nicholas Joseph, Greg Brockman, et~al.
\newblock Evaluating large language models trained on code.
\newblock {\em arXiv preprint arXiv:2107.03374}, 2021.

\bibitem{austin2021program}
Jacob Austin, Augustus Odena, Maxwell Nye, Maarten Bosma, Henryk Michalewski, David Dohan, Ellen Jiang, Carrie Cai, Michael Terry, Quoc Le, et~al.
\newblock Program synthesis with large language models.
\newblock {\em arXiv preprint arXiv:2108.07732}, 2021.

\bibitem{hendrycks2021measuring}
Dan Hendrycks, Collin Burns, Saurav Kadavath, Akul Arora, Steven Basart, Eric Tang, Dawn Song, and Jacob Steinhardt.
\newblock Measuring mathematical problem solving with the math dataset.
\newblock {\em arXiv preprint arXiv:2103.03874}, 2021.

\bibitem{cobbe2021training}
Karl Cobbe, Vineet Kosaraju, Mohammad Bavarian, Mark Chen, Heewoo Jun, Lukasz Kaiser, Matthias Plappert, Jerry Tworek, Jacob Hilton, Reiichiro Nakano, et~al.
\newblock Training verifiers to solve math word problems.
\newblock {\em arXiv preprint arXiv:2110.14168}, 2021.

\bibitem{zhang2024generativeverifiersrewardmodeling}
Lunjun Zhang, Arian Hosseini, Hritik Bansal, Mehran Kazemi, Aviral Kumar, and Rishabh Agarwal.
\newblock Generative verifiers: Reward modeling as next-token prediction, 2024.

\bibitem{wang2023selfconsistency}
Xuezhi Wang, Jason Wei, Dale Schuurmans, Quoc~V Le, Ed~H. Chi, Sharan Narang, Aakanksha Chowdhery, and Denny Zhou.
\newblock Self-consistency improves chain of thought reasoning in language models.
\newblock In {\em The Eleventh International Conference on Learning Representations}, 2023.

\bibitem{zheng2023judgingllmasajudgemtbenchchatbot}
Lianmin Zheng, Wei-Lin Chiang, Ying Sheng, Siyuan Zhuang, Zhanghao Wu, Yonghao Zhuang, Zi~Lin, Zhuohan Li, Dacheng Li, Eric~P. Xing, Hao Zhang, Joseph~E. Gonzalez, and Ion Stoica.
\newblock Judging llm-as-a-judge with mt-bench and chatbot arena, 2023.

\bibitem{lightman2023let}
Hunter Lightman, Vineet Kosaraju, Yura Burda, Harri Edwards, Bowen Baker, Teddy Lee, Jan Leike, John Schulman, Ilya Sutskever, and Karl Cobbe.
\newblock Let's verify step by step.
\newblock {\em arXiv preprint arXiv:2305.20050}, 2023.

\bibitem{liang2024improving}
Zhenwen Liang, Ye~Liu, Tong Niu, Xiangliang Zhang, Yingbo Zhou, and Semih Yavuz.
\newblock Improving llm reasoning through scaling inference computation with collaborative verification.
\newblock {\em arXiv preprint arXiv:2410.05318}, 2024.

\bibitem{rafailov2024direct}
Rafael Rafailov, Archit Sharma, Eric Mitchell, Christopher~D Manning, Stefano Ermon, and Chelsea Finn.
\newblock Direct preference optimization: Your language model is secretly a reward model.
\newblock {\em Advances in Neural Information Processing Systems}, 36, 2024.

\bibitem{ahmadian2024back}
Arash Ahmadian, Chris Cremer, Matthias Gall{\'e}, Marzieh Fadaee, Julia Kreutzer, Olivier Pietquin, Ahmet {\"U}st{\"u}n, and Sara Hooker.
\newblock Back to basics: Revisiting reinforce style optimization for learning from human feedback in llms.
\newblock {\em arXiv preprint arXiv:2402.14740}, 2024.

\bibitem{hao2023reasoning}
Shibo Hao, Yi~Gu, Haodi Ma, Joshua Hong, Zhen Wang, Daisy Wang, and Zhiting Hu.
\newblock Reasoning with language model is planning with world model.
\newblock In {\em Empirical Methods in Natural Language Processing}, pages 8154--8173, 2023.

\bibitem{xie2024self}
Yuxi Xie, Kenji Kawaguchi, Yiran Zhao, James~Xu Zhao, Min-Yen Kan, Junxian He, and Michael Xie.
\newblock Self-evaluation guided beam search for reasoning.
\newblock {\em Advances in Neural Information Processing Systems}, 36, 2024.

\bibitem{light2024scattered}
Jonathan Light, Yue Wu, Yiyou Sun, Wenchao Yu, Xujiang Zhao, Ziniu Hu, Haifeng Chen, Wei Cheng, et~al.
\newblock Scattered forest search: Smarter code space exploration with llms.
\newblock {\em arXiv preprint arXiv:2411.05010}, 2024.

\bibitem{chen2022codet}
Bei Chen, Fengji Zhang, Anh Nguyen, Daoguang Zan, Zeqi Lin, Jian-Guang Lou, and Weizhu Chen.
\newblock Codet: Code generation with generated tests.
\newblock {\em arXiv preprint arXiv:2207.10397}, 2022.

\bibitem{mcaleese2024llm}
Nat McAleese, Rai~Michael Pokorny, Juan Felipe~Ceron Uribe, Evgenia Nitishinskaya, Maja Trebacz, and Jan Leike.
\newblock Llm critics help catch llm bugs.
\newblock {\em arXiv preprint arXiv:2407.00215}, 2024.

\bibitem{gu2024survey}
Jiawei Gu, Xuhui Jiang, Zhichao Shi, Hexiang Tan, Xuehao Zhai, Chengjin Xu, Wei Li, Yinghan Shen, Shengjie Ma, Honghao Liu, et~al.
\newblock A survey on llm-as-a-judge.
\newblock {\em arXiv preprint arXiv:2411.15594}, 2024.

\bibitem{bigelow2024forking}
Eric Bigelow, Ari Holtzman, Hidenori Tanaka, and Tomer Ullman.
\newblock Forking paths in neural text generation.
\newblock {\em arXiv preprint arXiv:2412.07961}, 2024.

\bibitem{muennighoff2025s1}
Niklas Muennighoff, Zitong Yang, Weijia Shi, Xiang~Lisa Li, Li~Fei-Fei, Hannaneh Hajishirzi, Luke Zettlemoyer, Percy Liang, Emmanuel Cand{\`e}s, and Tatsunori Hashimoto.
\newblock s1: Simple test-time scaling.
\newblock {\em arXiv preprint arXiv:2501.19393}, 2025.

\bibitem{hendrycks2021measuring-MATH}
Dan Hendrycks, Collin Burns, Saurav Kadavath, Akul Arora, Steven Basart, Eric Tang, Dawn Song, and Jacob Steinhardt.
\newblock Measuring mathematical problem solving with the math dataset.
\newblock {\em arXiv preprint arXiv:2103.03874}, 2021.

\bibitem{xiong2024rlhflowmath}
Wei Xiong, Hanning Zhang, Nan Jiang, and Tong Zhang.
\newblock An implementation of generative prm.
\newblock \url{https://github.com/RLHFlow/RLHF-Reward-Modeling}, 2024.

\bibitem{wang2024math}
Peiyi Wang, Lei Li, Zhihong Shao, Runxin Xu, Damai Dai, Yifei Li, Deli Chen, Yu~Wu, and Zhifang Sui.
\newblock Math-shepherd: Verify and reinforce llms step-by-step without human annotations.
\newblock In {\em Proceedings of the 62nd Annual Meeting of the Association for Computational Linguistics (Volume 1: Long Papers)}, pages 9426--9439, 2024.

\bibitem{jain2024livecodebench}
Naman Jain, King Han, Alex Gu, Wen-Ding Li, Fanjia Yan, Tianjun Zhang, Sida Wang, Armando Solar-Lezama, Koushik Sen, and Ion Stoica.
\newblock Livecodebench: Holistic and contamination free evaluation of large language models for code.
\newblock {\em arXiv preprint arXiv:2403.07974}, 2024.

\bibitem{chen2025towards}
Qiguang Chen, Libo Qin, Jinhao Liu, Dengyun Peng, Jiannan Guan, Peng Wang, Mengkang Hu, Yuhang Zhou, Te~Gao, and Wanxiang Che.
\newblock Towards reasoning era: A survey of long chain-of-thought for reasoning large language models.
\newblock {\em arXiv preprint arXiv:2503.09567}, 2025.

\bibitem{chen2023teaching}
Xinyun Chen, Maxwell Lin, Nathanael Sch{\"a}rli, and Denny Zhou.
\newblock Teaching large language models to self-debug.
\newblock {\em arXiv preprint arXiv:2304.05128}, 2023.

\bibitem{zhou2023language}
Andy Zhou, Kai Yan, Michal Shlapentokh-Rothman, Haohan Wang, and Yu-Xiong Wang.
\newblock Language agent tree search unifies reasoning acting and planning in language models.
\newblock {\em ICML}, 2024.

\bibitem{wang2024planning}
Evan Wang, Federico Cassano, Catherine Wu, Yunfeng Bai, Will Song, Vaskar Nath, Ziwen Han, Sean Hendryx, Summer Yue, and Hugh Zhang.
\newblock Planning in natural language improves llm search for code generation.
\newblock {\em arXiv preprint arXiv:2409.03733}, 2024.

\bibitem{manvi2024adaptive}
Rohin Manvi, Anikait Singh, and Stefano Ermon.
\newblock Adaptive inference-time compute: Llms can predict if they can do better, even mid-generation.
\newblock {\em arXiv preprint arXiv:2410.02725}, 2024.

\bibitem{leea2025evolving}
Kuang-Huei Leea, Ian Fischera, Yueh-Hua Wuc, Dave Marwood, Shumeet Baluja, Dale Schuurmans, and Xinyun Chen.
\newblock Evolving deeper llm thinking.
\newblock {\em Gen}, 2:3, 2025.

\bibitem{guan2025rstarmathsmallllmsmaster}
Xinyu Guan, Li~Lyna Zhang, Yifei Liu, Ning Shang, Youran Sun, Yi~Zhu, Fan Yang, and Mao Yang.
\newblock rstar-math: Small llms can master math reasoning with self-evolved deep thinking, 2025.

\bibitem{chen2024think}
Xingyu Chen, Jiahao Xu, Tian Liang, Zhiwei He, Jianhui Pang, Dian Yu, Linfeng Song, Qiuzhi Liu, Mengfei Zhou, Zhuosheng Zhang, Rui Wang, Zhaopeng Tu, Haitao Mi, and Dong Yu.
\newblock Do not think that much for 2+3=? on the overthinking of o1-like llms, 2024.

\bibitem{wei2022chain}
Jason Wei, Xuezhi Wang, Dale Schuurmans, Maarten Bosma, Fei Xia, Ed~Chi, Quoc~V Le, Denny Zhou, et~al.
\newblock Chain-of-thought prompting elicits reasoning in large language models.
\newblock {\em NeurIPS}, 35:24824--24837, 2022.

\bibitem{sprague2024cotcot}
Zayne Sprague, Fangcong Yin, Juan~Diego Rodriguez, Dongwei Jiang, Manya Wadhwa, Prasann Singhal, Xinyu Zhao, Xi~Ye, Kyle Mahowald, and Greg Durrett.
\newblock To cot or not to cot? chain-of-thought helps mainly on math and symbolic reasoning, 2024.

\bibitem{wang2024chainofthoughtr}
Xuezhi Wang and Denny Zhou.
\newblock Chain-of-thought reasoning without prompting, 2024.

\bibitem{kojima2022large}
Takeshi Kojima, Shixiang~Shane Gu, Machel Reid, Yutaka Matsuo, and Yusuke Iwasawa.
\newblock Large language models are zero-shot reasoners.
\newblock {\em Advances in neural information processing systems}, 35:22199--22213, 2022.

\bibitem{zhouleast}
Denny Zhou, Nathanael Sch{\"a}rli, Le~Hou, Jason Wei, Nathan Scales, Xuezhi Wang, Dale Schuurmans, Claire Cui, Olivier Bousquet, Quoc~V Le, et~al.
\newblock Least-to-most prompting enables complex reasoning in large language models.
\newblock In {\em International Conference on Learning Representations}, 2023.

\bibitem{wangself}
Xuezhi Wang, Jason Wei, Dale Schuurmans, Quoc~V Le, Ed~H Chi, Sharan Narang, Aakanksha Chowdhery, and Denny Zhou.
\newblock Self-consistency improves chain of thought reasoning in language models.
\newblock In {\em International Conference on Learning Representations}, 2023.

\bibitem{li2023making}
Yifei Li, Zeqi Lin, Shizhuo Zhang, Qiang Fu, Bei Chen, Jian-Guang Lou, and Weizhu Chen.
\newblock Making language models better reasoners with step-aware verifier.
\newblock In {\em Proceedings of the 61st Annual Meeting of the Association for Computational Linguistics (Volume 1: Long Papers)}, pages 5315--5333, 2023.

\bibitem{alabdulmohsin2025recursive}
Ibrahim Alabdulmohsin and Xiaohua Zhai.
\newblock Recursive inference scaling: A winning path to scalable inference in language and multimodal systems.
\newblock {\em arXiv preprint arXiv:2502.07503}, 2025.

\bibitem{wang2025dynscaling}
Fei Wang, Xingchen Wan, Ruoxi Sun, Jiefeng Chen, and Sercan~{\"O} Ar{\i}k.
\newblock Dynscaling: Efficient verifier-free inference scaling via dynamic and integrated sampling.
\newblock {\em arXiv preprint arXiv:2506.16043}, 2025.

\bibitem{liventsev2023fully}
Vadim Liventsev, Anastasiia Grishina, Aki H{\"a}rm{\"a}, and Leon Moonen.
\newblock Fully autonomous programming with large language models.
\newblock In {\em Proceedings of the Genetic and Evolutionary Computation Conference}, pages 1146--1155, 2023.

\bibitem{chen2023evoprompting}
Angelica Chen, David Dohan, and David So.
\newblock Evoprompting: Language models for code-level neural architecture search.
\newblock {\em Advances in neural information processing systems}, 36:7787--7817, 2023.

\bibitem{romera2024mathematical}
Bernardino Romera-Paredes, Mohammadamin Barekatain, Alexander Novikov, Matej Balog, M~Pawan Kumar, Emilien Dupont, Francisco~JR Ruiz, Jordan~S Ellenberg, Pengming Wang, Omar Fawzi, et~al.
\newblock Mathematical discoveries from program search with large language models.
\newblock {\em Nature}, 625(7995):468--475, 2024.

\bibitem{lehman2023evolution}
Joel Lehman, Jonathan Gordon, Shawn Jain, Kamal Ndousse, Cathy Yeh, and Kenneth~O Stanley.
\newblock Evolution through large models.
\newblock In {\em Handbook of Evolutionary Machine Learning}, pages 331--366. Springer, 2023.

\bibitem{hemberg2024evolving}
Erik Hemberg, Stephen Moskal, and Una-May O’Reilly.
\newblock Evolving code with a large language model.
\newblock {\em Genetic Programming and Evolvable Machines}, 25(2):21, 2024.

\bibitem{chen2024divide}
Jingchang Chen, Hongxuan Tang, Zheng Chu, Qianglong Chen, Zekun Wang, Ming Liu, and Bing Qin.
\newblock Divide-and-conquer meets consensus: Unleashing the power of functions in code generation.
\newblock {\em arXiv preprint arXiv:2405.20092}, 2024.

\bibitem{zenkner2024abstractbeam}
Janis Zenkner, Lukas Dierkes, Tobias Sesterhenn, and Chrisitan Bartelt.
\newblock Abstractbeam: Enhancing bottom-up program synthesis using library learning.
\newblock {\em arXiv preprint arXiv:2405.17514}, 2024.

\bibitem{levin2025effective}
Kyla~H Levin, Kyle Gwilt, Emery~D Berger, and Stephen~N Freund.
\newblock Effective llm-driven code generation with pythoness.
\newblock {\em arXiv preprint arXiv:2501.02138}, 2025.

\bibitem{hernandez2recursive}
Sergio Hern{\'a}ndez-Guti{\'e}rrez, Minttu Alakuijala, Alexander~V Nikitin, and Pekka Marttinen.
\newblock Recursive decomposition with dependencies for generic divide-and-conquer reasoning.
\newblock In {\em The First Workshop on System-2 Reasoning at Scale, NeurIPS'24}, 2024.

\bibitem{khot2022decomposed}
Tushar Khot, Harsh Trivedi, Matthew Finlayson, Yao Fu, Kyle Richardson, Peter Clark, and Ashish Sabharwal.
\newblock Decomposed prompting: A modular approach for solving complex tasks.
\newblock {\em arXiv preprint arXiv:2210.02406}, 2022.

\bibitem{dua2022successive}
Dheeru Dua, Shivanshu Gupta, Sameer Singh, and Matt Gardner.
\newblock Successive prompting for decomposing complex questions.
\newblock {\em arXiv preprint arXiv:2212.04092}, 2022.

\bibitem{cazenave2009nested}
Tristan Cazenave.
\newblock Nested monte-carlo search.
\newblock In {\em Twenty-First International Joint Conference on Artificial Intelligence}, 2009.

\bibitem{cicirello2005max}
Vincent~A Cicirello and Stephen~F Smith.
\newblock The max k-armed bandit: A new model of exploration applied to search heuristic selection.
\newblock In {\em The Proceedings of the Twentieth National Conference on Artificial Intelligence}, volume~3, pages 1355--1361, 2005.

\bibitem{carpentier2014extreme}
Alexandra Carpentier and Michal Valko.
\newblock Extreme bandits.
\newblock {\em Advances in Neural Information Processing Systems}, 27, 2014.

\end{thebibliography}
